\documentclass[10pt,journal,compsoc]{IEEEtran}

\ifCLASSOPTIONcompsoc
  \usepackage[nocompress]{cite}
\else
  \usepackage{cite}
\fi
\ifCLASSINFOpdf
  \usepackage[pdftex]{graphicx}
\else
  \usepackage[dvips]{graphicx}
\fi

\usepackage{url}

\usepackage{tikz}
\usepackage{tikz-3dplot}
\usetikzlibrary{matrix, arrows, fit, calc, automata, shapes.gates.logic.US,shapes.gates.logic.IEC,calc, 
positioning, shapes.geometric, decorations.markings, decorations.pathreplacing}

\usepackage{subcaption}
\usepackage{booktabs}
\usepackage{amsmath}
\usepackage{multirow}
\usepackage[colorlinks,linkcolor=black, urlcolor=black, citecolor=black]{hyperref}

\newcommand{\bftab}{\fontseries{b}\selectfont}

\begin{document}
%
\title{ABMOF: A Novel Optical Flow Algorithm for Dynamic Vision Sensors}

\author{Min~Liu,~\IEEEmembership{Member,~IEEE,}
        and~Tobi~Delbruck,~\IEEEmembership{Fellow,~IEEE}
\IEEEcompsocitemizethanks{\IEEEcompsocthanksitem M. Liu and T. Delbruck are with Sensors Group, affiliated with the Institute
of Neuroinformatics, University of Zurich and ETH Zurich, Switzerland: \href{http://sensors.ini.uzh.ch}{http://sensors.ini.uzh.ch}. 
Corresponding author E-mail: minliu@ini.uzh.ch.
\protect\\
This research is funded by Swiss National Center of Competence in Research Robotics (NCCR Robotics). We thank iniLabs for providing technical support, and L. Jira and T. Zaluska for help with recordings.
}
}

\markboth{ABMOF: A Novel Optical Flow Algorithm for Dynamic Vision Sensors}%
{Shell \MakeLowercase{\textit{et al.}}: Bare Advanced Demo of IEEEtran.cls for IEEE Computer Society Journals}
\IEEEtitleabstractindextext{%
\begin{abstract}
Dynamic Vision Sensors (\textbf{DVS}), which output asynchronous log intensity change events, have potential applications in high-speed robotics, autonomous cars and drones. The precise event timing, sparse output, and wide dynamic range of the events are well suited for optical flow, but conventional optical flow (\textbf{OF}) algorithms are not well matched to the event stream data. This paper proposes an event-driven OF algorithm called adaptive block-matching optical flow (\textbf{ABMOF}). ABMOF uses time slices of accumulated DVS events. The time slices are adaptively rotated based on the input events and OF results.  Compared with other methods such as gradient-based OF, ABMOF can efficiently be implemented in compact logic circuits. Results show that ABMOF achieves comparable accuracy to conventional standards such as Lucas-Kanade (\textbf{LK}). The main contributions of our paper are new adaptive time-slice rotation methods that ensure the generated slices have sufficient features for matching,including a feedback mechanism that controls the generated slices to have average slice displacement within the block search range. An LK method using our adapted slices is also implemented. The ABMOF accuracy is compared with this LK method on natural scene data including sparse and dense texture, high dynamic range, and fast motion exceeding 30,000 pixels per second. The paper dataset and source code are available from \href{http://sensors.ini.uzh.ch/databases.html}{http://sensors.ini.uzh.ch/databases.html}.
\end{abstract}

\begin{IEEEkeywords}
dynamic vision sensor, optical flow, event-based sensor, block matching, neuromorphic.
\end{IEEEkeywords}}

\maketitle

\IEEEdisplaynontitleabstractindextext

\IEEEpeerreviewmaketitle

\IEEEraisesectionheading{\section*{Supplementary Material}}


\begin{itemize}
    \item \href{https://drive.google.com/open?id=10X0z4zznuV9j1OOjWpJGv-YCWujkF7FiYjG6efwUrP0}{\underline{Video and ABMOF18 dataset}}.
    \item  \href{https://github.com/SensorsINI/jaer/tree/ivy-bagreader/src/ch/unizh/ini/jaer/projects/minliu}{ch.unizh.ini.jaer.projects.minliu} source code
\end{itemize}

\section{Introduction}

\label{sec:introduction}

\IEEEPARstart{C}{omputing} optical flow (OF) is a fundamental problem of computer vision. 
There are a variety of algorithms for frame-based cameras. 
The most widely used method in computer vision
is probably the efficient and highly optimized LK method~\cite{lucas1981iterative} from OpenCV~\cite{opencv_optical_flow}.
It is a sparse method that locally applies brightness constancy around detected feature points. Most recent developments have been used deep convolutional neural networks to compute dense flow; they achieve high accuracy but are extremely expensive, for example FlowNet2.0 \cite{ilg_flownet_2016} runs HD video at 8\,FPS using several hundred watts of PC+GPU power. 

Although LK is widely used, it fails
when the images are over or underexposed, and when the images are
too blurred to extract good features, and when these features have too much displacement
between frames, as we demonstrate in Sec.~\ref{sec:results}. 
These scenarios arise in high speed vision, 
for example in fast drone flight, or under low lighting conditions, 
where the frame exposure increases, or under natural lighting conditions, 
where extreme lighting variations, glare, and lens flare are common.

The adaptive block matching optical flow~(\textbf{ABMOF}) OF method proposes to address these problems. It is 
a semi-dense method that computes flow at points where brightness changes.
These brightness changes come from a Dynamic Vision Sensor
(\textbf{DVS}) silicon retina. The DVS is a new type of 
camera~\cite{lichtsteiner_128_2008,ebnes,delbruckISCAS,brandli_240x180_2014,li_rgbw_2015,posch_asynchronous_2008} that
provides sparse asynchronous data output, high dynamic range,  
high time resolution and low latency. 
Its output is a variable 
data-rate stream of timestamped pixel brightness change events.
The DVS requires new algorithms to take advantage of these brightness change events.
The DVS has been proposed for high-level vision algorithms such as 
localization~\cite{weikersdorfer2012event}, 
navigation~\cite{mueggler2015continuous, gallego2016event}, 
landing~\cite{de2013optic}\cite{hordijk2017vertical}, 
visual odometry~\cite{zhu_event-based_2017, kueng_low-latency_2016, censi_low-latency_2014}, and simultaneous localization and mapping~\cite{weikersdorfer2013simultaneous}.
Previous low-level algorithms for DVS OF are reviewed in Sec.~\ref{sec:priorof}.

The ABMOF algorithm originates from video compression motion estimation.
The core of the algorithm is an event-driven computation of block matching 
optical flow that operates on variable duration time slice images of accumulated DVS events.
ABMOF is directly targeted for 
efficient multiplier-free parallel hardware implementation using simple logic circuits.
Our original BMOF~\cite{liu2017block} does 
not work well on complex scenes, but we here describe five improvements that form ABMOF:

\begin{itemize}
 \item A new \textsl{AreaEventCount} slice rotation method 
 correctly rotates slices that vary in density of features.
\item A new feedback control mechanism for adapting slice duration 
 achieves a target average match distance, increasing speed range and usability.
 \item A new adaptive event-skipping mechanism does not discard any sensor data but only computes OF when pipeline allows it.
\item Using new multiscale time slices matches longer distances and lower spatial frequencies. 
\item Using a diamond search rather than full search improves search efficiency, e.g.
by about 14X for search distances of 12 pixels.

\end{itemize}

Comparing both block matching and Lucas-Kanade methods on the adaptive slices shows improvement for both methods compared with the previous fixed slice durations.

\subsection{DVS and DAVIS}
DVS pixels independently react to brightness (log intensity) changes.
If any pixel detects a brightness increase or decrease that exceeds a critical threshold amount,
relative to the previously-memorized brightness, 
it generates an output event, and memorizes the new brightness value. 
Each event consists of a timestamp with microsecond resolution, 
an event address represented by x and y pixel location, and a polarity, where 1 means ON event increase 
and 0 means OFF event decrease. 
Comparing with a conventional camera, the DVS thus has worse spatial resolution but better temporal resolution. It also has lower latency and higher dynamic range.

The DAVIS~\cite{brandli_240x180_2014} combines the DVS with conventional Active Pixel Sensor (\textbf{APS}) 
technology in the same pixel, using a shared photodiode. 
The DAVIS concurrently outputs DVS events and gray-scale image frames.
While the DVS has high dynamic range~(\textbf{DB}) (typically $>$120dB), the APS output has only limited DR of about 55dB. We use the APS output here to compare ABMOF with conventional LK on the frames.

\subsection{Prior DVS Optical Flow}
\label{sec:priorof}

This section reviews previous DVS OF algorithms. They have been dominated by event-driven methods that operate on each DVS event.

\cite{delbruck2008frame} described an open-source algorithm (called \textbf{DS} in this paper) for time-of-flight DVS OF based on oriented edges detected by spatio-temporal coincidence. It works only for sharp edges and suffers from aperture problems since it is edge-based.

\cite{Benosman2012} adapted the frame-based LK algorithm (called \textbf{EBLK} in this paper) to DVS.
It stores a fixed-queue length window of past events. 
For each new event, it computes the LK algorithm on a window of fixed time interval of a block of pixels 
surrounding the current event pixel.
The gradient estimation precision is low due to quantization and small 5x5 window size. The small window size was used to limit the computation time in order to keep up with a high rate of events.

\cite{barranco2014contour} proposed a contour-based method.
In their work, they compared events-only method and events-frames-combined method. The difference is that by using events-only sensor, they need to reconstruct the contrast of the edge to localize the contour but it is not necessary for frames which have the absolute intensity. 
The optical flow estimation then is obtained from the contour width divided by the time interval.

\cite{benosman2014event} proposed a 
time-surface method (called \textbf{LP} in this paper) that combines the 
2D events and timestamps into 3D space. Normal OF is
obtained by robust iterative local plane fitting.
It works well for sharp edges but fails with dense textures, thin lines, and natural scenes~\cite{rueckauer2016evaluation,zhu_ev-flownet:_2018}
since these both produce complex structures that plane fitting does not model.

\cite{barranco2015bio} proposed a 
more expensive phase-based method for high-frequency texture regions.
They use normalized cross-correlation of to measure the pixel's 
timestamps' similarity and localize the contour. 
Once the contour is found, they use a Gabor filter to extract the local phase. 
The OF constraint that assumes the 
constancy of the spatio-temporal contours using the phase
is formulated and is used to solve the normal OF flow. 

\cite{rueckauer2016evaluation} implemented and
compared the DS, EBLK, and LP methods. 
It concluded that the existing algorithms were both 
computationally expensive and do not work well natural scenes and noisy sensor data.
This paper also proposed an evaluation method 
and a provided a simple benchmark dataset with ground truth.
The ground truth OF is obtained by constraining 
the camera motion to pure rotation and uses 
the camera's Inertial Measurement Unit~(\textbf{IMU}) 
rate gyros to obtain global translational and rotational OF.

\cite{bardow2016simultaneous} proposed 
a frame-based variational algorithm that simultaneously estimates 
the optical flow, gradient map, and intensity 
reconstruction from DVS. 
Although the simultaneous constraints  
results in very regularized output, the results are not quantified, 
and the method is very expensive compared to others.

Though the main goal of \cite{zhu2017event} 
is for event-based feature tracking, 
it also proposed a pipeline to compute OF on corner points. 
They added another two assumptions: 
One is that events generated by the same point lie on a curve, 
and OF within a small spatial temporal window is constant.
The OF problem is cast in an optimization 
framework and the expectation maximization (EM) algorithm computes 
the solution. It can run in real time with 15 features on a PC.

\cite{zhu_ev-flownet:_2018} recently 
reported the first CNN-based DAVIS OF architecture and published a useful dataset. 
It is trained by minimizing photometric loss from the DAVIS APS frames.
It achieves the best published accuracy, but burns 
50W to run at 25\,Hz frame rate 
on a laptop gamer GPU.

The above methods are serial algorithms that solve linear or nonlinear constraints; 
some of them use iteration or exclusion of events 
to make the solutions more robust. 
All methods so far have required at least several us/event 
on a fast PC that consumes many tens of watts.  

{\bf Hardware implementations:} \cite{conradt2015board} developed a microcontroller-based embedded implementation of a time of flight (TOF) OF method. 
It works well for isolated points but not for dense textured scenes; 
it also has aperture problems with edges. 
A simplified version of \cite{benosman2014event} using 
 3x3 windows was implemented on FPGA in~\cite{tun_aung_event-based_2018}. The small windows restricts it to sharp edges.

This paper extends on our previous FPGA BMOF implementation~\cite{liu2017block}. We used a method called block-matching. 
Block-matching was developed for motion 
estimation for MPEG video encoders and there are many silicon implementations.  
We demonstrated only a basic implementation for a small block size of 9x9 pixels
using a single pixel shift 
in the NSEW compass directions and a single fixed time slice duration.
It does not work well on most real scenes. 
But the advantage of block matching is that in hardware, 
large blocks can be matched with few clock cycles and simple logic. 
For example, 21x21 blocks can easily be implemented by parallel logic 
circuits, and these large block sizes are important for good flow accuracy with noisy sensor data.

Here, as described in this paper, we have extended from the original implementation.
 The paper is organized as follows: Sec.~\ref{sec:approach} explains the idea of block-matching 
and our improvements. Sec.~\ref{sec:results} shows experimental results, and Sec.~\ref{sec:conclusion} concludes the paper.

\section{ABMOF algorithm}
\label{sec:approach}

\begin{figure}[h!]

 \centering
    \tikzstyle{branch}=[fill,shape=circle,minimum size=3pt,inner sep=0pt]
    \tikzstyle{startstop} = [rectangle, rounded corners,
                            minimum width=2cm, minimum height=1cm,
                            text centered, draw=black, fill=red!30]
    \tikzstyle{io} = [trapezium, trapezium left angle=70, trapezium right angle=110,
                      minimum width=2cm, minimum height=1cm, inner sep=10pt,
                      text centered, draw=black, fill=blue!30]
    \tikzstyle{process} = [rectangle, minimum width=2cm, minimum height=1cm,
                           text centered, text width=2cm, draw=black, fill=orange!30]
    \tikzstyle{decision} = [diamond, minimum width=2cm, minimum height=1cm, aspect=2, inner sep=1pt,
                           text centered, draw=black, fill=green!30]
    \tikzstyle{arrow} = [thick,->,>=stealth]
    \tikzstyle{output} = [rectangle, minimum width=1cm, minimum height=1cm,
                           text centered, text width=1cm, draw=black, fill=red!30]
        \begin{tikzpicture}[label distance = 4mm, scale = 1, node distance = 2cm]
            \node (start) [startstop] {Input DVS event};
            \node (rotation logic) [decision, below of=start, xshift=0cm, yshift=0cm] {Rotate slice?};
            \node (accumulation) [process, below of=rotation logic, yshift=-0.5cm, inner sep=5pt] {Accumulate events to multi-scale slices};
            \node (skipping) [decision, below of=accumulation, yshift=-3.5cm] {Skip current event?};

            \node (OF cal) [process, right of=skipping, xshift=2cm] {OF calculation};
            \node (outliers) [process, above of=OF cal, yshift=-0.2cm] {Remove outliers};
            \node (feedback) [process, above of=outliers, yshift=-0.2cm] {Update OF histogram};
            \node (output)  [output, left of = outliers, xshift = -0.5cm] {OF \\output};
            \node (adaptation) [process, above of=feedback, yshift=-0.2cm] {Adapt slice duration};

            \draw [arrow] (start) -- (rotation logic);
            \draw [arrow] (rotation logic) -- node[anchor=east] {no} (accumulation);
            \draw [arrow] (rotation logic) -| node[anchor=south, xshift=-1.5cm] {yes} (adaptation);

            \draw [arrow] (accumulation) -- (skipping);
            
            \draw [arrow] (skipping) -- node[anchor=south] {yes}([xshift=-1cm]skipping.west) |- (start); 
            \draw [arrow] (skipping) -- node[anchor=south] {no} (OF cal);
            \draw [arrow] (OF cal) -- (outliers);
            \draw [arrow] (outliers) -- (feedback);
            \draw [arrow] (outliers) -- (output);
            \draw [arrow, dashed] (feedback) -- (adaptation);
            \draw [arrow] (adaptation) -- ([xshift=0.3cm]adaptation.east) |- ([yshift=-0.4cm]start.south);             
            \draw [arrow] (adaptation) -- (accumulation);

        \end{tikzpicture}
    \caption{The pipeline of our algorithm}\label{fig:framework_alg}
\end{figure}
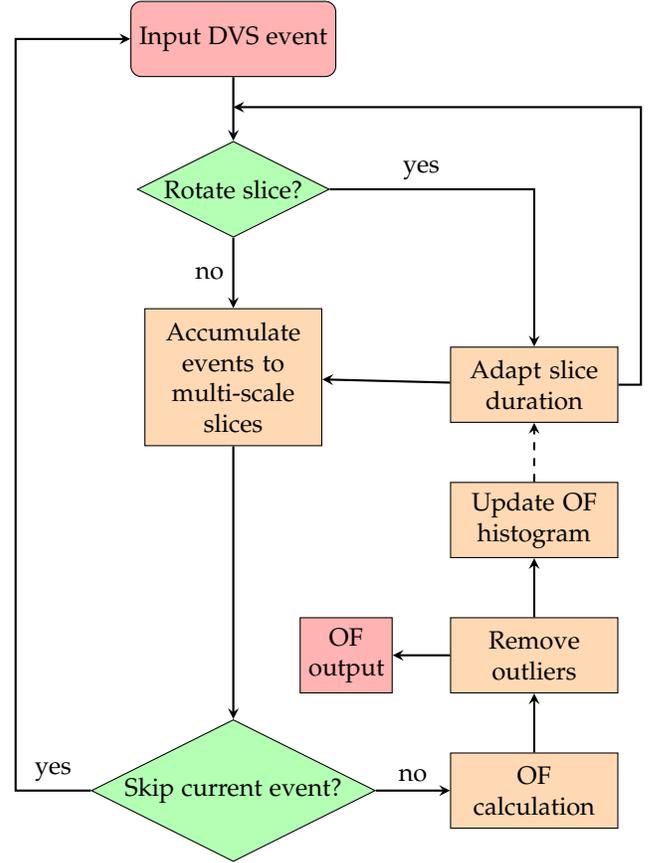

The pipeline of ABMOF is summarized in Fig~\ref{fig:framework_alg} and the time slices and block matching are illustrated in Fig.~\ref{fig:BMOF_basic}.
When a new event arrives, 
the event's timestamp is used by the rotation logic to 
determine whether the event slice is to be rotated. 
If yes, the slices are rotated and the slice duration $d$ or event count parameter $K$ or $k$
is adapted based on the current slices's OF distribution.
The adapted slice duration is sent as an input to the rotation logic.
The adaptation takes the OF distribution of the previous slice
as the input.
We use a dashed connection in the figure to represent their relationship.
Details of the rotation logic are introduced in Sec.~\ref{Slice Rotation}. 

All the new events will be accumulated to multi-scale slices. 
If the system is busy, the OF calculation for the event is skipped.
Otherwise it triggers the OF calculation. 
The event skipping mechanism is introduced in Sec.~\ref{event_skipping}.
After removing outliers, the OF histogram is updated. 

All the parameters that are used in this paper are summarized in Table~\ref{tab:parameters}. 

\begin{table}[!htbp]
    \caption{Symbols, description, and typical values/units.}\label{tab:parameters}
        \begin{center}
            \begin{tabular}{c c c}
                \toprule
                \bftab symbol & \bftab description & \bftab typical values (default)\\
                \midrule
                $w\times h$ & width $\times$ height of pixel array  &  346x260 \\
               $d$ &  slice duration & 1\textendash100\,ms (50)\\
                $K$ &  global event number & 1k\textendash50k\,events (10k)  \\
                $k$ &  area event number & 100\textendash1k\,events (1k)\\
                $a$ &  area dimension subsampling & 5 bits \\
                $b$ & block dimension & 11\textendash21\,pixels (21)\\
                $r$ &  search radius & 4\textendash12\,pixels (4)\\
                $s$ &  \# scales & 1\textendash3 (2)\\
                $p$ &  skip count on PC for real-time & 30\textendash1000 \\
                $g$ & \# bits for slice counts  & 1\textendash7 (3)\\
               $g$ & \# bits for slice counts  & 1\textendash7 (3)\\
                $D$  &  average match distance & ideally $r/2$\\     
                $(v_x, v_y)$  &  OF result & pixels/sec (pps) \\         
                \bottomrule  
            \end{tabular}
        \end{center}
\end{table}

\subsection{Block-Matching DVS Time Slices}
\begin{figure}[!htbp]
\begin{center}
  \includegraphics[width=9cm]{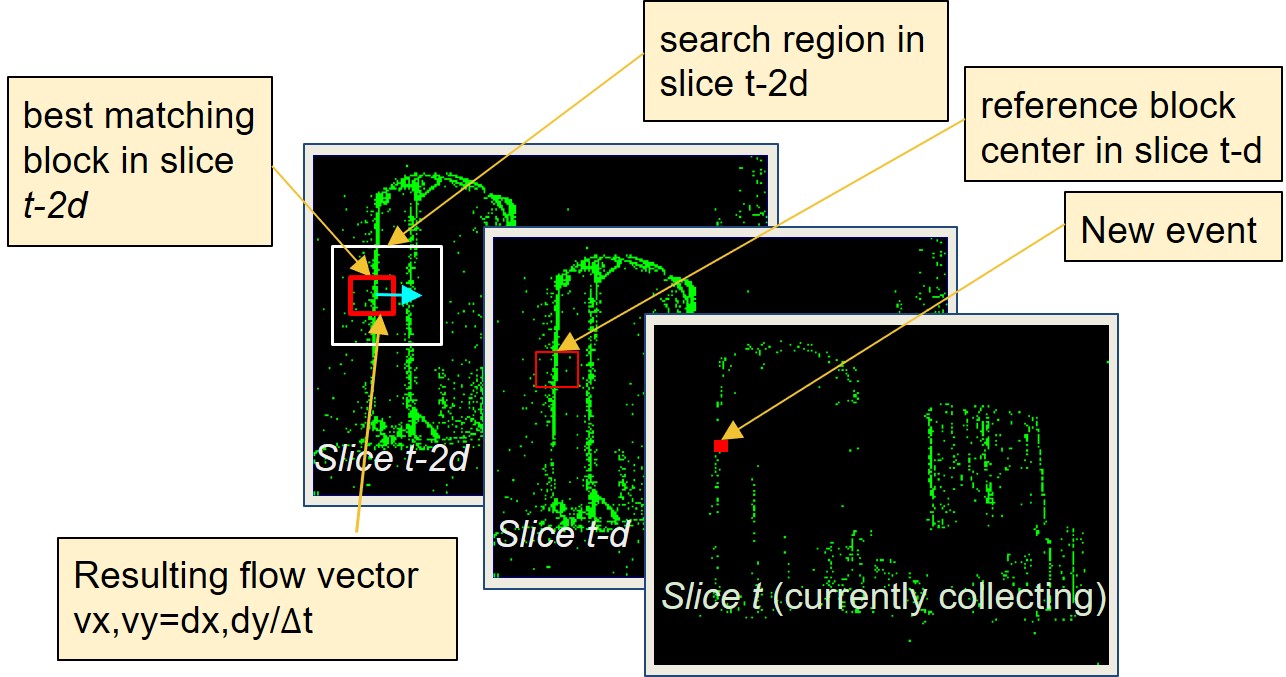}
\caption{BMOF block matching, on \texttt{boxes} from \cite{rueckauer2016evaluation}}\label{fig:BMOF_basic}
\end{center}
\end{figure}

Fig.~\ref{fig:BMOF_basic} shows the main principle of BMOF:
Three time-slice memories store the events as 2D event histograms:  Slice \textit{t} accumulates the current events. 
Slices \textit{t-d} and \textit{t-2d} hold the previous two slices.  
\textit{d} is the slice duration.
When a new event arrives, 
it is accumulated to slice \textit{t}
by either incrementing the pixel value,
or adding the polarity of the event to it.
Which of these is done depends on if we
ignore the event polarity. For the experiments
in this paper, we usually ignored event 
polarity because the accuracy did not change significantly when we included it.
Including polarity may enable better block matching,
but it carries the price that one bit of the pixel memory
is used for the sign bit.
After accumulating the event, then the other two slices
are then used to compute the OF based on the current
event's location. When multi-scale slices
are used (Sec.~\ref{sec:multiscale}), then each slice is a pyramid of $s$ slices.

A reference block ($b*b$ pixels) is centered 
on the incoming event's 
location on the \textit{t-d} slice map (red box in slice \textit{t-d}).
The best matching block 
on the \textit{t-2d} slice is found based on Sum of Absolute Difference~(\textbf{SAD}) 
inside a $(2r+1)^2$ search region, shown as a white rectangle in the  \textit{t-2d} slice.
Thus the optical flow result is obtained by using these two blocks' offset $(d_x,d_y)$, 
divided by the time interval $\Delta t$ between these two slices. 
The time of each slice is taken as the average 
of the first and last timestamp of events accumulated to each slice.

The slices are rotated according to slice rotation logic (Sec.~\ref{Slice Rotation}).
The rotation discards the \textit{t-2d} slice and uses its memory for the new slice \textit{t}; similarly slice \textit{t} becomes slice \textit{t-d} and slice \textit{t-d} becomes slice \textit{t-2d}.
In \cite{liu2017block}, the slice duration \textit{d} was set by user manually. 
It is not convenient for general application since limits the speed range. 
In this paper, we propose several methods to adjust it adaptively.

\subsection{Slice Rotation Methods} \label{Slice Rotation}
Slice rotation is the core part of our algorithm.
It calculates when to rotate the slices to ensure good slice quality.
Good slices should have sharp features, not too much displacement, and not be too sparse.
This goal is achieved by feed-forward and feedback control.
We show the details of these two algorithms in the following subsections.

\subsubsection{Feedforward Slice Rotation}

The new events are accumulated into the latest slice, slice \textit{t}.
Slice \textit{t} is only used for accumulation.
After that, it will be rotated to be as a past slice and used for OF calculation. 

In our original BMOF work~\cite{liu2017block}, we implemented the method \textsl{ConstantDuration}, where each slice has the same duration $d$. Another obvious method is to rotate slices after a constant number $K$ of events have been accumulated, called \textsl{ConstantEventNumber}. 
\begin{itemize}
\item \textsl{ConstantDuration}: Here the slices are accumulated to time slices uniformly with duration $d$.
    This method is what we reported before and corresponds most closely to conventional frame based methods.
        It has the disadvantage that if the scene motion is too fast, 
        then the movement between slices may be too large to be matched using a specified search distance.
        If the movement is too slow, then the features may not move enough between slices, 
        resulting in reduced flow speed and angle resolution.
\item \textsl{ConstantEventNumber}: Here the slices are accumulated until they contain a fixed total count of DVS events $K$. 
    If $K$ is large then the slices will tend to have larger $d$.
        But if the scene moves faster, then the rate of DVS events also increases, 
        which for fixed $K$ will decrease $d$.
        Therefore the \textsl{ConstantEventNumber} method automatically adapts $d$ to the average overall scene dynamics.
\end{itemize}

A drawback of the \textsl{ConstantEventNumber} method is its global nature.
For scenes which have lots of or very few textures, it is impossible to set a suitable global $K$. In order to address this problem, we propose a new rotation method called \textsl{AreaEventNumber}.

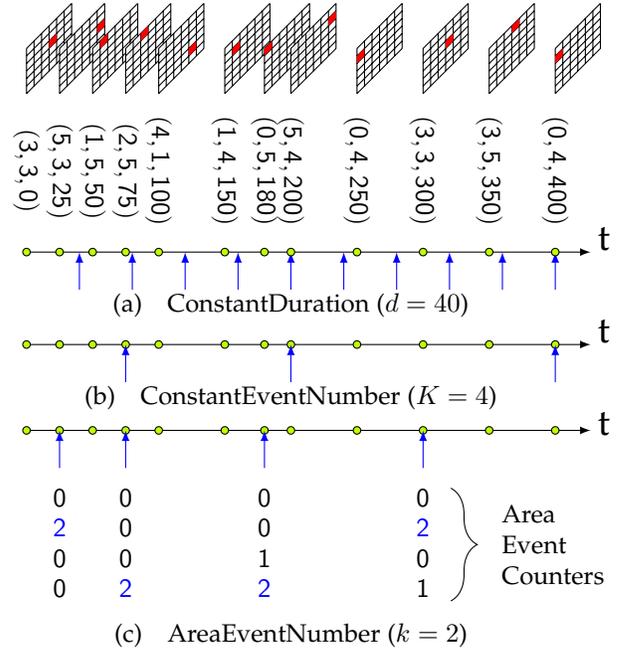
\begin{figure}[!htbp]
\begin{tikzpicture}[scale=.5,every node/.style={minimum size=1cm}]
    
    \def \gridWidth{0.2};
    \def \areaFlg{0};
    \newcounter{areaCntA} 
    \setcounter{areaCntA}{0}
    \newcounter{areaCntB} 
    \setcounter{areaCntB}{0}
    \newcounter{areaCntC} 
    \setcounter{areaCntC}{0}
    \newcounter{areaCntD} 
    \setcounter{areaCntD}{0}


    \draw[-latex, yshift=-120] (0,0) -- (15,0);
    \node[xshift=5pt, yshift=-55] at(15,0) {\Large t};
    \node[xshift=100pt, yshift=-80] at(0,0) {(a)\quad   ConstantDuration ($d=40$)};

    \draw[-latex, yshift=-190] (0,0) -- (15,0);
    \node[xshift=5pt, yshift=-90] at(15,0) {\Large t};
    \node[xshift=100pt, yshift=-115] at(0,0) {(b)\quad   ConstantEventNumber ($K=4$)};

    \draw[-latex, yshift=-255] (0,0) -- (15,0);
    \node[xshift=5pt, yshift=-125] at(15,0) {\Large t};
    \node[xshift=100pt, yshift=-205] at(0,0) {(c)\quad   AreaEventNumber ($k=2$)};
    \draw [decorate,decoration={brace,amplitude=10pt,raise=4pt}]
            (11,-10.5) -- (11,-13.5) node [black,midway,xshift=1.5cm,align=left] {Area \\ Event \\ Counters}; 

  \pgfmathsetseed{3}
    \foreach[count=\i] \xshift in {0,25,...,80,100,150,180,200,250,...,400}
  {
	\pgfmathsetmacro{\x}{int(random(0,5))};
	\pgfmathsetmacro{\y}{int(random(0,5))};
    \pgfmathsetmacro{\timeCnt}{int(\xshift/40)};
    \pgfmathsetmacro{\eventCnt}{int(mod(\i, 4))};

    \begin{scope}[
            yshift=0, xshift=\xshift, every node/.append style={
     	    yslant=0.5,xslant=-1},yslant=1,xslant=0
            ]
        \fill[white,fill opacity=0.9] (0,0) rectangle (1.2,1.2);
        \draw[step=\gridWidth, black] (0,0) grid (1.2,1.2); 
        \fill[red] (\gridWidth*\x,\gridWidth*\y) rectangle (\gridWidth*\x+\gridWidth,\gridWidth*\y+\gridWidth);
    \end{scope}

    \begin{scope}[
    		yshift=-10, xshift=\xshift
            ]
        \node [black, rotate=270] at(0, -1.7) {$\mathsf{(\x, \y, \xshift)}$};       

    \end{scope}

    \draw [fill=lime, yshift=-120, xshift=\xshift] (0,0) circle (.1) ;
    \ifthenelse{\i<11} {
        \begin{scope}[
                yshift=-120, xshift=\i*40
            ]
            \draw [blue, latex-] (0,0) to[in=90, out=270] (0, -1) ;
        \end{scope}
    }{}

    \draw [fill=lime, yshift=-190, xshift=\xshift] (0,0) circle (.1) ;
    \ifnum\eventCnt=0
        \begin{scope}[
                yshift=-190, xshift=\xshift
            ]
            \draw [blue, latex-] (0,0) to[in=90, out=270] (0, -1) ;
        \end{scope}
    \fi

    \draw [fill=lime, yshift=-255, xshift=\xshift] (0,0) circle (.1) ;
    \ifthenelse{\x<3 \AND \y<3 \AND \x>-1 \AND \y>-1}{
    	\addtocounter{areaCntA}{1}
    }{}
    \ifthenelse{\x<6 \AND \y<6 \AND \x>2 \AND \y>2}{
    	\addtocounter{areaCntB}{1}
    }{}
    \ifthenelse{\x<6 \AND \y<3 \AND \x>2 \AND \y>-1}{
    	\addtocounter{areaCntC}{1}
    }{}
    \ifthenelse{\x<3 \AND \y<6 \AND \x>-1 \AND \y>2}{
    	\addtocounter{areaCntD}{1}
    }{}
    
    \ifthenelse{\theareaCntA>1 \OR \theareaCntB>1 \OR \theareaCntC>1 \OR \theareaCntD>1} {
        \pgfmathsetmacro{\areaFlg}{int(1)};
    }{}

	\ifthenelse{\areaFlg=1} {
        \begin{scope}[
                yshift=-255, xshift=\xshift
            ]
            \draw [blue, latex-] (0,0) to[in=90, out=270] (0, -1) ;

            \ifthenelse{\theareaCntA=2}{
                \node [blue, rectangle] at(0, -1.8) {$\mathsf{\theareaCntA}$}; 
            }{
                \node [black, rectangle] at(0, -1.8) {$\mathsf{\theareaCntA}$}; 
            }
            \ifthenelse{\theareaCntB=2}{
                \node [blue, rectangle] at(0, -2.6) {$\mathsf{\theareaCntB}$};       
            }{
                \node [black, rectangle] at(0, -2.6) {$\mathsf{\theareaCntB}$};       
            }
            \ifthenelse{\theareaCntC=2}{
                \node [blue, rectangle] at(0, -3.4) {$\mathsf{\theareaCntC}$};       
            }{
                \node [black, rectangle] at(0, -3.4) {$\mathsf{\theareaCntC}$};       
            }
            \ifthenelse{\theareaCntD=2}{
                \node [blue, rectangle] at(0, -4.2) {$\mathsf{\theareaCntD}$};    
            }{
                \node [black, rectangle] at(0, -4.2) {$\mathsf{\theareaCntD}$};    
            }

            \setcounter{areaCntA}{0}
            \setcounter{areaCntB}{0}
            \setcounter{areaCntC}{0}
            \setcounter{areaCntD}{0}
        \end{scope}
    }{}

    }

\end{tikzpicture}
    \caption{Three feedforward slice rotation methods. The event stream is at the top of the figure.
    The information including event address (x,y) and timestamp is shown under the event stream.
    An example of these three slice methods is demonstrated here, with (a) slice duration $d=40$,
    (b) global event number $K=4$ and (c) area event number $k=2$.}\label{fig:AreaEventNumber}
\end{figure}

\begin{itemize}
\item \textsl{AreaEventNumber}: Instead of rotating the slices based on the sum of the whole slice event number, 
\textsl{AreaEventNumber} will trigger the slice rotation once any one of the area's event number 
(Area Event Counters) exceeds the threshold value $k$. Each area is $(w\times h)/2^a$\,pixels, i.e. $10\times 8$\,pixels.
\end{itemize}

By using the \textsl{AreaEventNumber} method, slice rotation is data-driven by the accumulation of DVS events,
but adapts the slice durations to match the area of the scene which has the most DVS activity.
This adaptation prevents under-sampling that causes displacement that is too large to match between slices.
Compared with \textsl{ConstantEventNumber} method, it preserves the advantage that the generated slices adapt to the scene dynamics.
The local adaptability makes the slices more robust to variation and distribution of scene texture. 

To make it even more robust and adaptive, 
the slice event number $k$ is also adaptive to the scene.
When the scene moves fast, the parameter $k$ will be increased. Otherwise, it is decreased. Adaptation of $k$ is further described in Sec.~\ref{feedback_rotation}.

An example to demonstrate these three methods is shown in Fig~\ref{fig:AreaEventNumber}. 
The blue arrows pointing to the three time axes represent these three rotation method results. 
It is obvious that both the time interval and the event number
interval are fixed for \textsl{ConstantDuration} and \textsl{ConstantEventNumber}. 
However, both of them vary in the \textsl{AreaEventNumber} 
method which makes it more adaptive to the dynamic scene. 

\begin{figure}[!htbp]
\begin{subfigure}[b]{0.2\textwidth}
        \begin{center}
            \includegraphics[width=4.2cm]{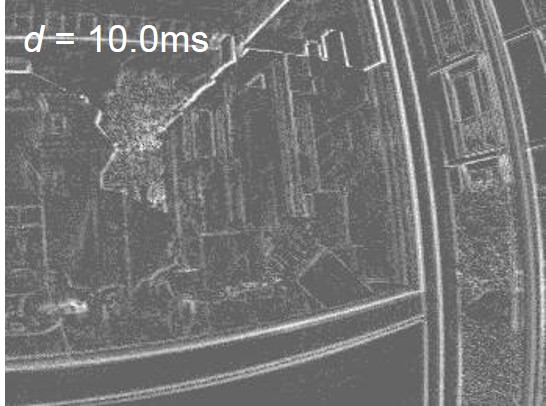}
        \end{center}
        \caption{\textsl{ConstantDuration} on dense texture scene}
        \label{fig:constantduration_result}
    \end{subfigure}    
    \hspace{1cm}
   \begin{subfigure}[b]{0.2\textwidth}
        \begin{center}
            \includegraphics[width=4.2cm]{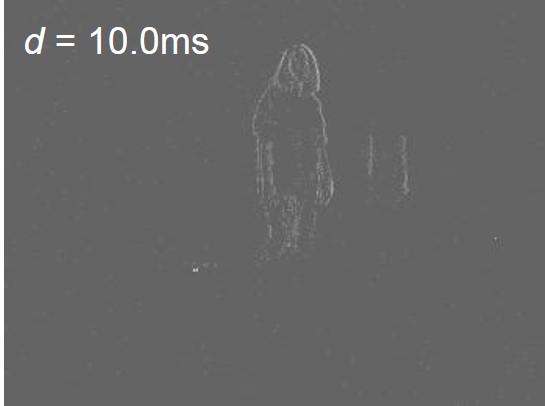}
        \end{center}
        \caption{\textsl{ConstantDuration} on sparse texture scene}
        \label{fig:constantduration_result_sparse}
    \end{subfigure}
    
    \begin{subfigure}[b]{0.2\textwidth}
        \begin{center}
           \includegraphics[width=4.2cm]{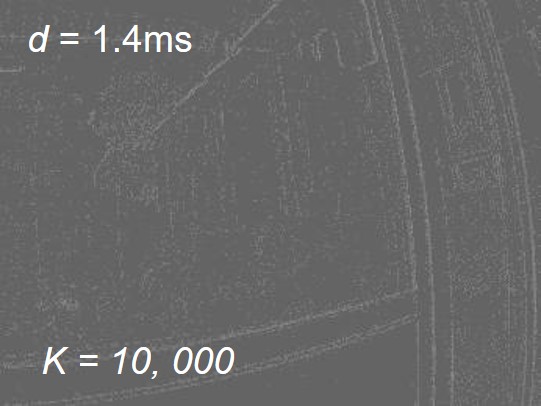}
        \end{center}
        \caption{\textsl{ConstanEventNumber} on dense texture scene}
        \label{fig:constanteventnubmer_result}
   \end{subfigure}
   \hspace{1cm}
   \begin{subfigure}[b]{0.2\textwidth}
        \begin{center}
            \includegraphics[width=4.2cm]{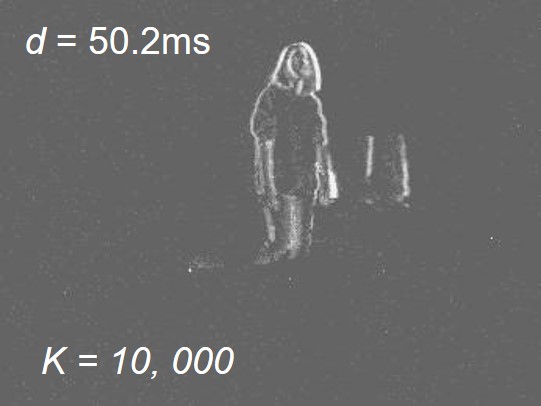}
        \end{center}
        \caption{\textsl{ConstanEventNumber} on sparse texture scene}
        \label{fig:constanteventnubmer_result_sparse}
    \end{subfigure}    
    
   \begin{subfigure}[b]{0.2\textwidth}
        \begin{center}
            \includegraphics[width=4.2cm]{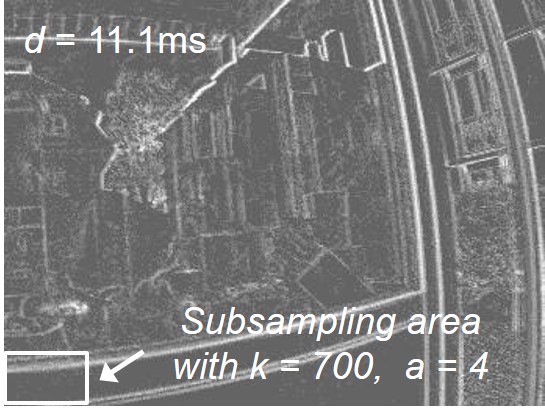}
        \end{center}
        \caption{\textsl{AreaEventNumber} on dense texture scene}
        \label{fig:areaeventnumber_result}
    \end{subfigure}
    \hspace{1cm}
    \begin{subfigure}[b]{0.2\textwidth}
        \begin{center}
           \includegraphics[width=4.2cm]{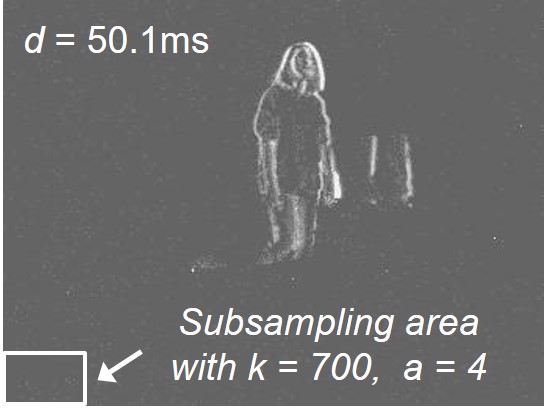}
        \end{center}
        \caption{\textsl{AreaEventNumber} on sparse texture scene}
        \label{fig:areaeventnumber_result_sparse}
   \end{subfigure}
    \caption{Comparison between event slices generated by three methods.
    }  
    \label{fig:Aread_and_Constant}
\end{figure}
In Fig~\ref{fig:Aread_and_Constant}, we compare these three methods on two different scenes, one have sparse textures and the other have dense textures. 
Among them, Figs.~\ref{fig:constantduration_result},~\ref{fig:constanteventnubmer_result}, and ~\ref{fig:areaeventnumber_result} are obtained from the same dense scene by the three different methods.  
Figs.~\ref{fig:constantduration_result_sparse},~\ref{fig:constanteventnubmer_result_sparse}, and ~\ref{fig:areaeventnumber_result_sparse} are obtained from a sparse scene. 
Both of the dense and sparse scenes use the same parameter for every method which is $d=10ms$ for \textsl{ConstantDuration}, $K=10000$ for \textsl{ConstantEventNumber} and $k=700$ for \textsl{AreaEventNumber}. 
The resulting slice durations are shown overlaid on each scene. 

Neither \textsl{ConstantDuration} nor  \textsl{ConstantEventNumber} work well on both of these
two scenes with fixed values of $d$ or $K$. 
For example, \textsl{ConstantDuration} fails in
the sparse scene because $d$ was set for the faster motion in
Fig.~\ref{fig:Aread_and_Constant}a and the duration
was too short for the slower motion in Fig.~\ref{fig:Aread_and_Constant}b.
\textsl{ConstantEventNumber} makes the slice too short in duration in the dense scene in Fig.~\ref{fig:Aread_and_Constant}c,
because $K$ was set to make a good slice for Fig.~\ref{fig:Aread_and_Constant}d.
However, \textsl{AreaEventNumber} with 
fixed parameter $k$ functions well on
both of scenes, because it correctly creates
the Fig.~\ref{fig:Aread_and_Constant}f slice
after being set for the dense scene in Fig.~\ref{fig:Aread_and_Constant}e. 
It shows that \textsl{AreaEventNumber} is more robust to dynamic scene content.


\subsubsection{Feedback Control of Slice Duration} \label{feedback_rotation}

Another method to automatically adjust the slice duration
is possible via feedback control. 
An optical flow distribution histogram is reset after each slice rotation and then collects the distribution of OF results. 
The histogram’s average match distance $D$ is calculated. 
If  $D>r/2$ where $r$ is the search radius, it means that the slice duration is too long, 
and so the slice duration or event number is decreased. 
Otherwise, if $D<r/2$, 
then it indicates the slices are too brief in duration, and the slice duration or event number is increased. 
It is possible that slice durations that are too brief or lengthy result in OF results of very small matching distance 
that are the result of a bias in the search algorithm towards zero motion (small match distance). 
Stability is improved by limiting the slice duration range within application-specific limits.
For the control policy, we so far used bang-bang control. 
A fixed factor of $\pm5\%$ adjusts the slice duration, 
where the sign of the relative change of duration 
is the sign of $r/2-D$. More sophisticated 
control policies are clearly possible,
since the value of the error 
is directly predictive of the necessary change in the duration.

\begin{figure}[!htbp]
\begin{center}
   \begin{subfigure}[b]{0.5\textwidth}
        \begin{center}
          \includegraphics[width=9cm]{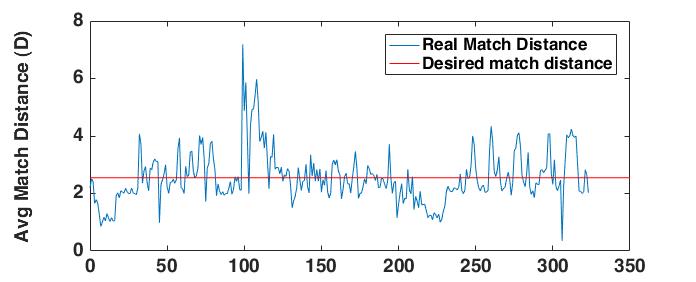}
        \end{center}
        \caption{OF result's match distance}
        \label{fig:feedback_a}
    \end{subfigure}

    \begin{subfigure}[b]{0.5\textwidth}
        \begin{center}
          \includegraphics[width=9cm]{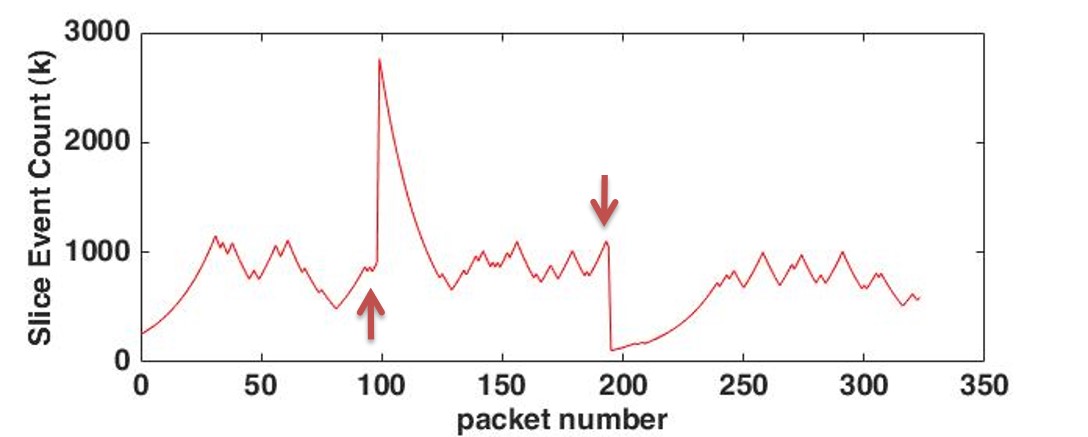}
        \end{center}
        \caption{Feedback on slice event count}
        \label{fig:feedback_b}
   \end{subfigure}
\caption{Feedback on slice event number. (a) shows the OF result's real match distance and its desired match distance. (b) represents the slice event count number.}\label{fig:feedback}
\end{center}
\end{figure}
Since the principle of the feedback control on event number and slice duration is similar, we  show only an example of feedback control on event number $k$ here. 
The data in Fig.~\ref{fig:feedback} shows an example of event number control using the \textsl{AreaEventNumber} rotation policy with feedback control of $k$. Fig.~\ref{fig:feedback_a} shows the average OF match distance $D$. The feedback control of event number holds $D$ at its $r/2$ value of about 2.5\,pixels. Fig.~\ref{fig:feedback_b} shows the event number $k$. It has a steady state value of about 1000. Around packet 100 (1st arrow), $k$ was manually perturbed to a large value, resulting in a increase in $D$, but it rapidly returns to the steady state value. At around packet 200 (2nd arrow), $k$ was manually reduced to a small value, resulting in a small $D$ of about 1\,pixel. Again $D$ returns to the steady state value. This data shows the stability of the event number control with this feedback mechanism.

\subsection{Search Method}
\label{sec:searchmethod}
The implementation of \cite{liu2017block} searched only the target block and its 8 nearest neighbors. 
An improvement is offered by extending the search range to a larger distance range $r$.
The block matching search method can be done by exhaustive full search. 
It has the best search accuracy but is expensive since the cost grows quadratically with $r$. 
A more efficient method is diamond search~\cite{zhu1997new}, which we implemented.
It makes a trade off between computation and accuracy. 
Our results shows that it has about 90\% chance to 
hit on the best matching block with a cost 14X less than the full search, for $r=12$. 
Using the diamond search improves the algorithm's real-time performance significantly.

\subsection{Multi Scale and Multi-Bit Time Slices}
\label{sec:multiscale}
A limitation of the approach described so far is the limited dynamic speed range of the method, 
since matching can only cover a spatial range of square radius $r$ around the reference location.
One way to increase the search range by a factor of $2^s$ with only $s$ linear increase in search time is to use a multi-scale pyramid~\cite{lindeberg1994scale}.
In this method, events are accumulated into a stack of time slices.
Each slice in the stack subsamples the original event addresses in x and y directions 
by a factor of 2 more than the previous scale.
I.e. if $s=0$ means the original full resolution scale, 
then events are accumulated into scale $s$ by first right shifting the event \textit{x} and \textit{y} addresses by $s$ bits,
and then accumulating the resulting event into the scale $s$ slice, 
which has only $1/2^s$ as many pixels for each dimension.
For example, in the $s=1$ scale slice, each pixel accumulates events from a 2x2 pixel region 
in the full resolution original pixel address space. To prevent saturation, we use multiple bits $g$ for each value; for example $g=3$ allows up to 7 unsigned events 
for each pixel when we ignore the event polarity, or up to $\pm 2$ events when we use polarity. 
Thus the total slice memory required for an 
$N$ pixel sensor is 
$3 N g \sum_{m=0}^{s-1}{2^{-m}}$ bits.

To compute the OF, each event is processed independently for each scale.
The match that has the minimum SAD is selected as the OF.
Using multiple scales is beneficial particularly in noisy situations, where the event flow is sparse. The binning of events helps to find good matches.

\subsection{Adaptive Event skipping} \label{event_skipping}
For high speed or densely textured scenes, the event rate becomes high.
If we still compute OF for each event the 
real-time performance will be influenced dramatically and the algorithm
quickly falls behind the actual incoming event rate.
To address this problem, we propose an event skipping method. 
If the processing time is higher than a threshold we set,
the following events do not have their OF calculated.
However, they are still accumulated to the current slice. 
By doing this, we can get a trade off between the OF density 
and the real-time performance. The adaptive event skipping algorithm 
uses a skip parameter $p$, which is increased or decreased depending 
on the frame rate. If the actual frame rate is slower than 
the desired frame rate set by the user, 
it means it takes too much time to process the event, then $p$ increases.
Otherwise, it decreases. 
On a Corei7-975 PC in Java 1.8, the ABMOF implementation requires about 15\,us per event with the default parameters in Table~\ref{tab:parameters} and $p=1$.
With $p=1000$, the time drops to an average of about 260\,ns/event; there is some overhead for each packet and for slice rotation, which is why it only drops by only a factor of 600X.
It is also easy to implement in hardware. A FIFO forms a buffer for the incoming events.
The event skipping will be designed as a switch and will be connected to the FIFO half-full flag. 
If the FIFO is half-full, it means the processing time 
is falling behind, and event skipping will be enabled.

\subsection{Outlier Rejection}

To improve the accuracy, we developed outlier rejection 
to filter out OF results with poor matching quality. We use two parameters to reject the outliers.
One parameter is \texttt{validPixOccupancy}; it determines 
the percentage of valid pixels in two blocks that will be compared.
Valid pixels are the pixels where events were accumulated. 
The reason for setting this parameter is sometimes the blocks 
are too sparse, which makes the distance metric get a meaningless result.
By only calculating matching for blocks that are filled
with sufficient valid pixels, we can reject misleading
results. 

A second outlier rejection parameter is called \texttt{maxAllowedSadDistance}.
The minimum distance 
between the reference block and the
candidate block must be smaller
than \texttt{maxAllowedSadDistance},
otherwise the OF event will 
be rejected.
Thus, the best matching search block may actually be a poor match, 
and \texttt{maxAllowedSadDistance} 
allows rejecting best matches if the match distance is too large.

\begin{figure}[!htbp]
\begin{subfigure}[b]{0.24\textwidth}
        \begin{center}
            \includegraphics[width=4cm]{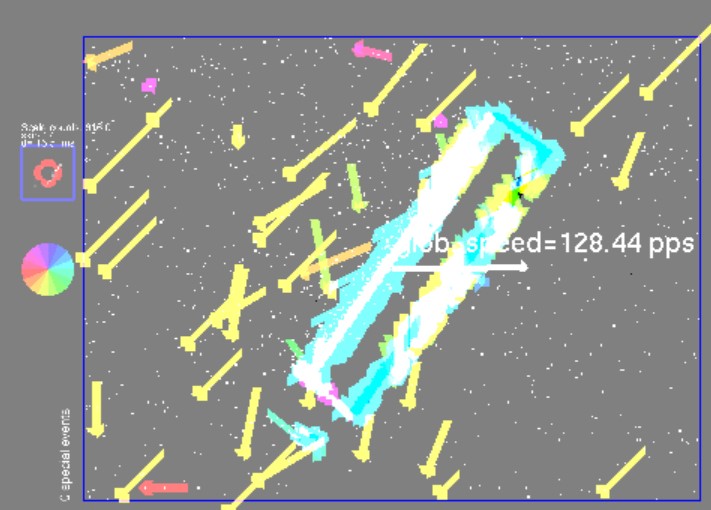}
        \end{center}
        \caption{Without outliers rejection}
        \label{fig:boxes}
    \end{subfigure}
    \begin{subfigure}[b]{0.24\textwidth}
        \begin{center}
           \includegraphics[width=4cm]{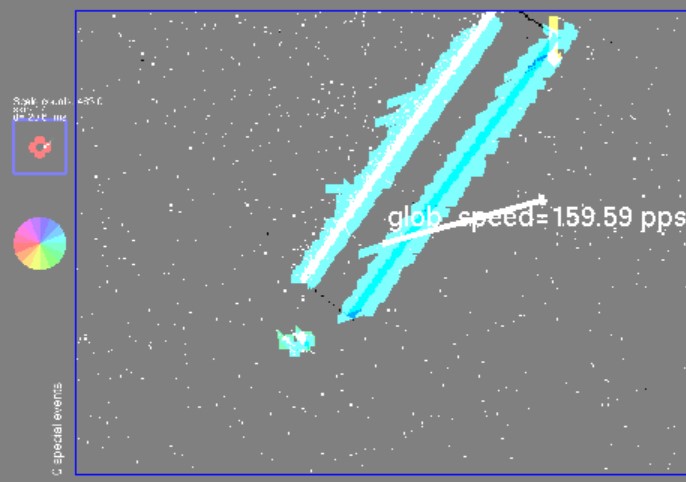}
        \end{center}
        \caption{With outliers rejection}
        \label{fig:grassPavement}
   \end{subfigure}
\caption{Example of outlier rejection using \texttt{maxAllowedSadDistance} and \texttt{validPixOccupancy}. (a): without outlier rejection. (b): using outlier rejection with \texttt{maxAllowedSadDistance=0.5} and \texttt{validPixOccupancy}=0.01.}\label{fig:outlier}
\end{figure}

The effect of these parameters is shown in example data in Fig~\ref{fig:outlier} from a simple case of a black bar moving up and to the the right. The flow results are visibly cleaner using these outlier rejection criteria.

Both outlier rejection mechanisms are easily implemented in hardware. For example, the valid pixel occupancy can be realized by pixel subtraction units that output a large value if both operands are zero. The confidence threshold can be realized by a comparator on the final best match output result that flags a distance that is too large.

\section{Results}
\label{sec:results}
The algorithms were implemented in jAER~\cite{JAER}.
In this section, we show several experiments to validate the ABMOF algorithm. 
They can be classified into two types.
The first type are the experiments with ground truth OF. 
We tested it on two datasets:  
\texttt{slider\_hdr\_far} 
is from \cite{mueggler2017event}. 
It shows a flat poster scene with fixed depth of 58\,cm where the camera 
is moved laterally by a motorized cart, resulting in uniform flow of about 90 pixels per second\,(pps).  
It was recorded with high lighting contrast.
\texttt{pavement\_fast} is a scene with extremely fast flow of 
34k\,pps recording from a car, with a down-looking camera recording an asphalt pavement.
The files converted to jAER aedat format 
are included in our dataset. 
We show both the qualitative and 
quantitative results in these experiments; see Sec.~\ref{Type I}

In the second type of experiment, we test our algorithm on several complex scenes. 
They consist of camera rotation over gravel (\texttt{gravel}), 
flow through an indoor office environment (\texttt{office}), 
and uniform flow created by a variety of shapes (\texttt{shapes}, from~\cite{mueggler2017event}).
The dataset is provided\footnote{\href{https://drive.google.com/open?id=10X0z4zznuV9j1OOjWpJGv-YCWujkF7FiYjG6efwUrP0}{ABMOF dataset README link for review}} 
to support the tests for other future algorithms. 
Due to unknown ground truth in these files, 
we show only a comparison between the ABMOF and 
Lukas Kanade results using the generated slices for these data; see Sec.~\ref{Type II}.

For the new data we gathered for this work, 
we used an unpublished advanced DAVIS with 346x260 pixel resolution and 
integrated on-chip APS readout circuits, allowing a maximum APS frame rate of about 50\,Hz. 
This DAVIS346 has pixels with integrated microlenses, 
optimized photodiodes, and antireflection coating, 
which together increase the effective quantum efficiency to about 24\%
compared with the previous DAVIS240C QE of 7\%.

For the \texttt{slider\_hdr\_far} data, 
the groundtruth camera position is provided for each time point and the scene has a provided uniform depth.
By using the camera calibration data and pinhole camera model, we converted the pose groundtruth data to a global optical flow groundtruth.

For \texttt{pavement\_fast}, we manually measured the flow using a jAER~\cite{JAER} software filter called \textsl{Speedometer}, which allows using the mouse to mark a moving feature point at different time points and measures the distance and time between these marks.

\subsection{Type I experiment result} \label{Type I}

We show the results of type I experiments in this subsection.
We measured four metrics to evaluate the algorithms. They are event density (\textbf{ED}), translational global flow (\textbf{GF}), 
Average Endpoint Error (\textbf{AEE}) and Average Angular Error (\textbf{AAE}). ED is the fraction of DVS events that result in OF results. 
DVS events are skipped because block matching fails to pass outlier rejection tests; we set $p=1$ for these tests. ED relates to the density of the flow computation. LK has very low density because it relies on features. An ED of 100\% means that all pixel brightness changes result in OF events. AEE and AAR are defined for DVS OF in~\cite{rueckauer2016evaluation}. 

Besides the ABMOF,  we also implemented the Lucas Kanade (\textbf{LK}) OF calculation based on our generated adaptive event slices using OpenCV and we call it ABMOF\_{LK}.
ABMOF\_{LK} uses our algorithm to set the time slice duration, 
and these generated slices are treated as conventional gray scale image frames. 
In ABMOF\_{LK}, corners are first extracted by Shi-Tomasi corner detector~\cite{shi1994good} and then they are are passed to the LK tracking algorithm implemented in OpenCV~\cite{opencv_optical_flow}.
LK estimates the OF result based on these features. 

We also compared the AMBOF OF methods with previously published implementations from~\cite{rueckauer2016evaluation}: \textsl{DirectionSelectiveFlow} (DS)~\cite{delbruck2008frame}, the event-based \textsl{LucasKanadeKFlow} (EBLK)~\cite{Benosman2012}, and \textsl{LocalPlanesFlow} (LP)~\cite{benosman2014event}.

\subsubsection{\texttt{slider} scene}
\begin{figure*}[!htpb]
	\begin{subfigure}[b]{0.33\textwidth}
        \begin{center}
            \includegraphics[width=6cm]{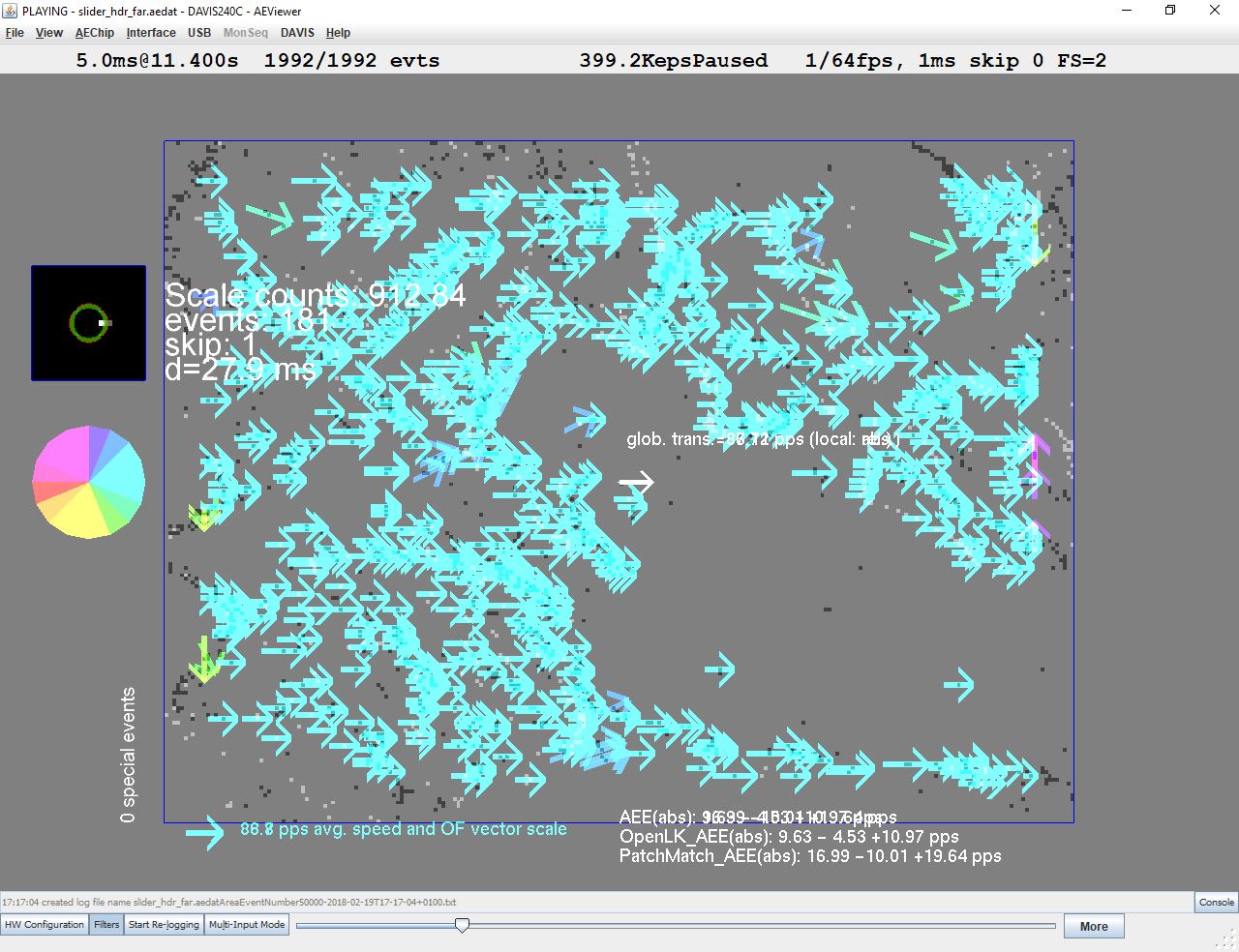}
        \end{center}
        \caption{\textrm{ABMOF}}
        \label{fig:slider_hdr_far_a}
    \end{subfigure}
	\begin{subfigure}[b]{0.33\textwidth}
        \begin{center}
            \includegraphics[width=6cm]{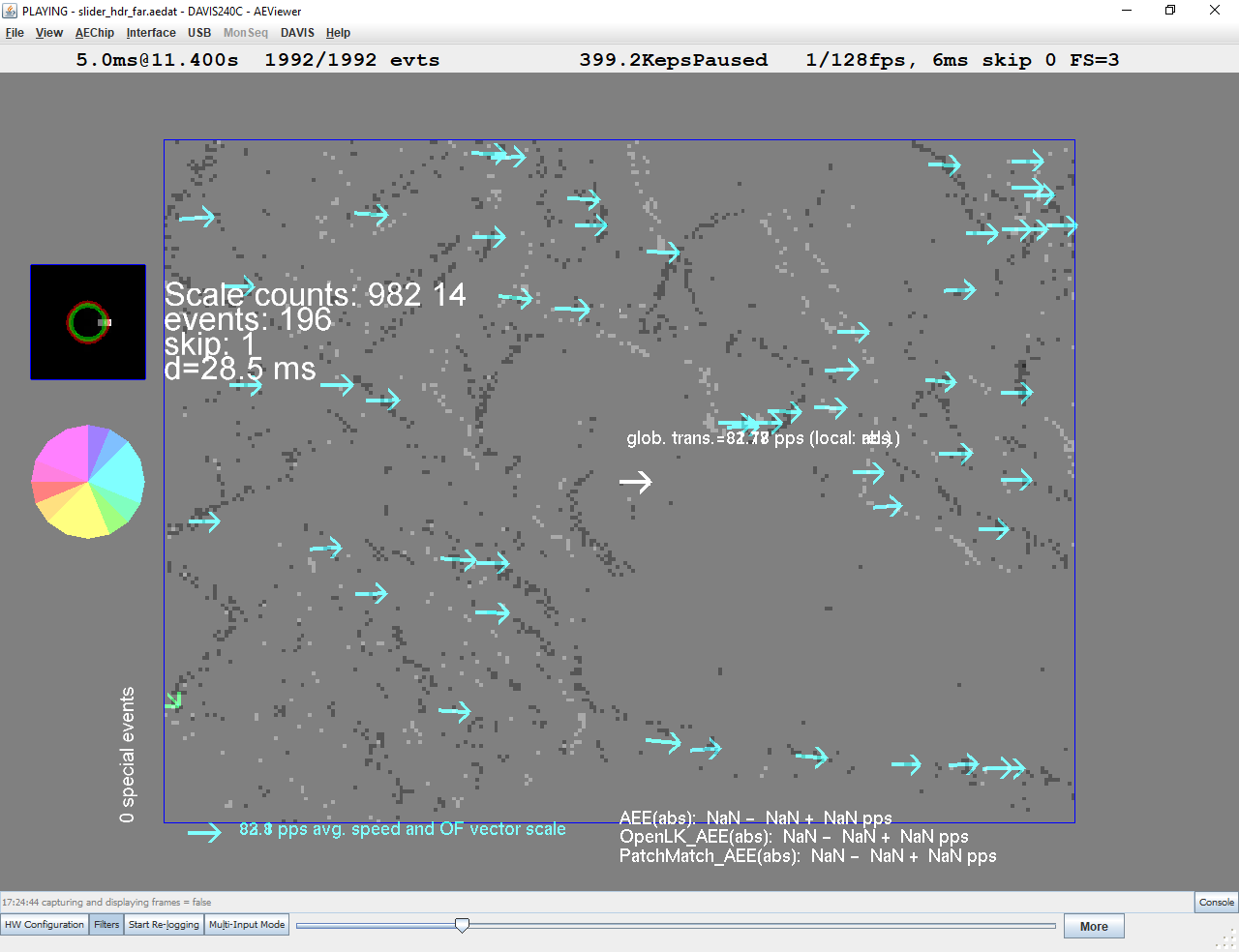}
        \end{center}
        \caption{ABMOF\_{LK}}
        \label{fig:slider_hdr_far_b}
    \end{subfigure}
    \begin{subfigure}[b]{0.33\textwidth}
        \begin{center}
            \includegraphics[width=6cm]{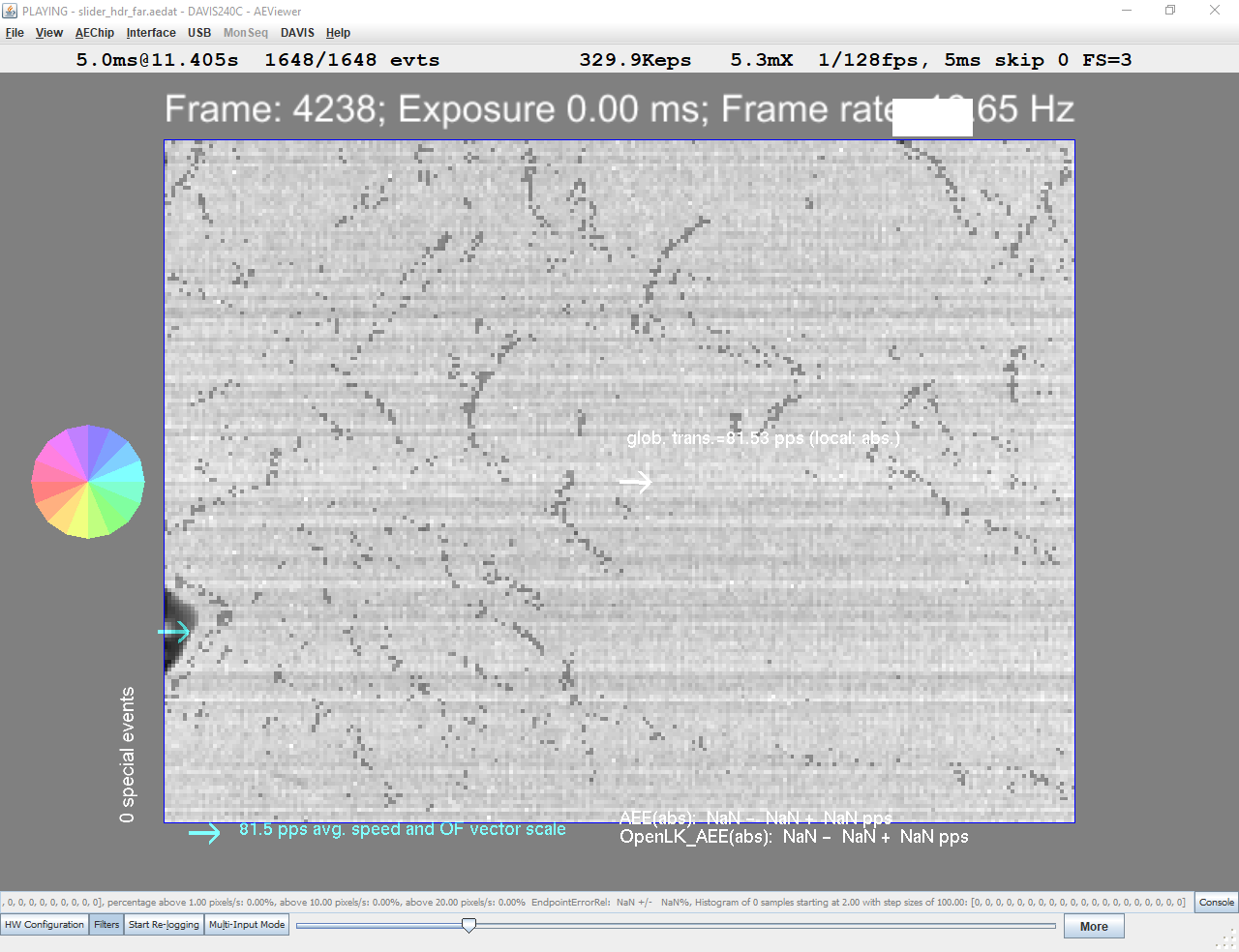}
        \end{center}
        \caption{APS image 2 and LK result}
        \label{fig:slider_hdr_far_c}
   \end{subfigure}
   	\begin{subfigure}[b]{0.33\textwidth}
        \begin{center}
            \includegraphics[width=6cm]{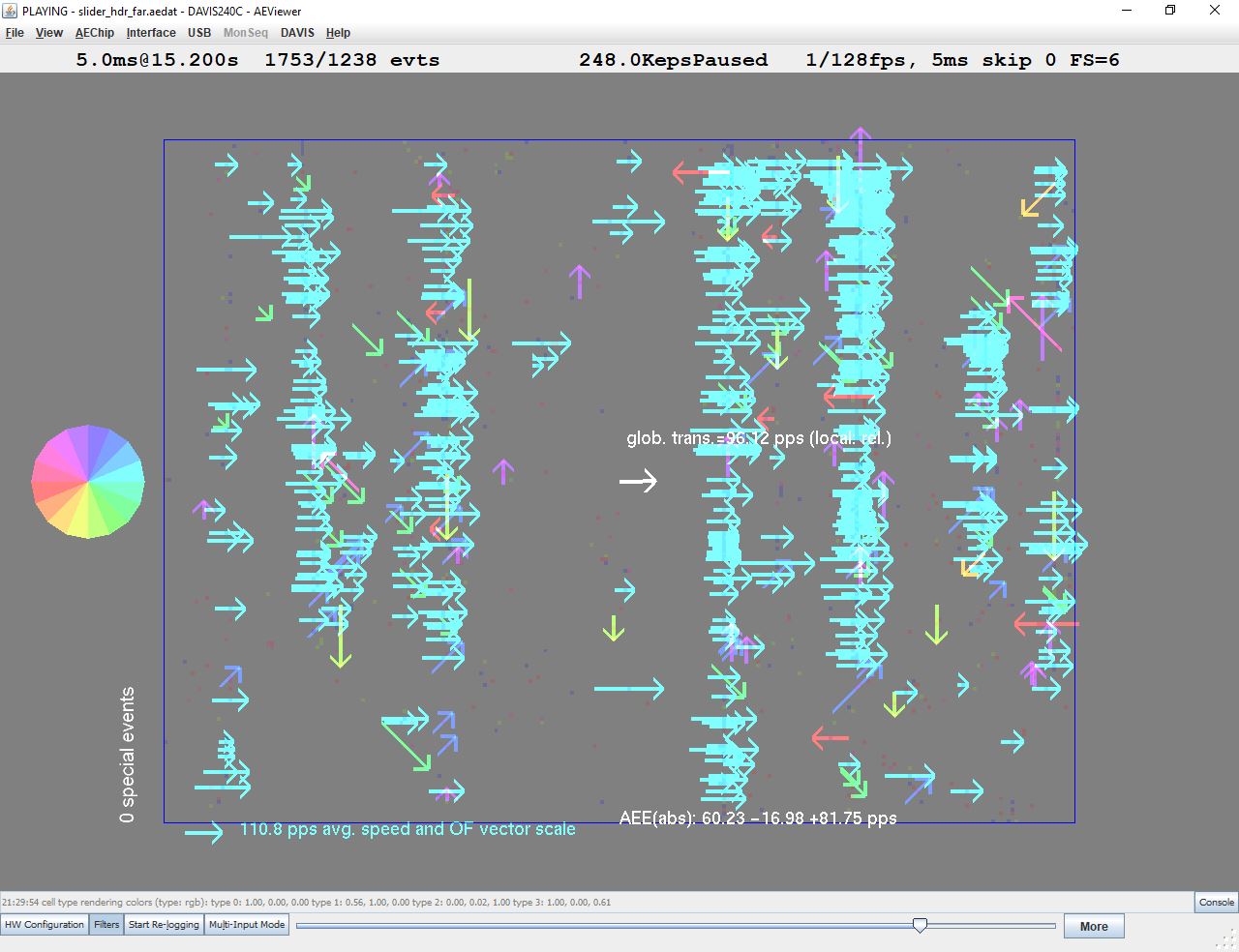}
        \end{center}
        \caption{\textrm{DS}}
        \label{fig:slider_hdr_far_d}
    \end{subfigure}
	\begin{subfigure}[b]{0.33\textwidth}
        \begin{center}
            \includegraphics[width=6cm]{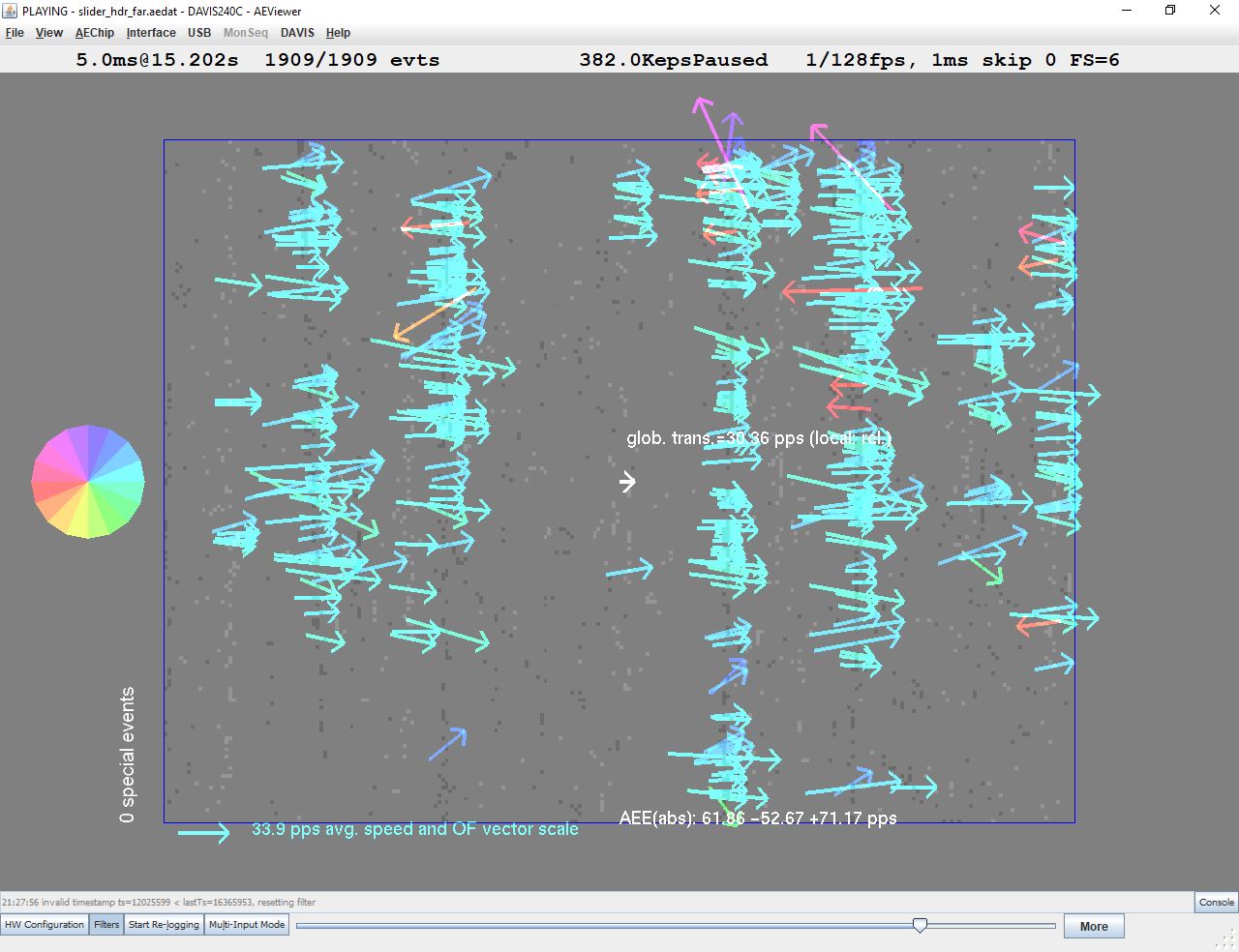}
        \end{center}
        \caption{EBLK}
        \label{fig:slider_hdr_far_e}
    \end{subfigure}
    \begin{subfigure}[b]{0.33\textwidth}
        \begin{center}
            \includegraphics[width=6cm]{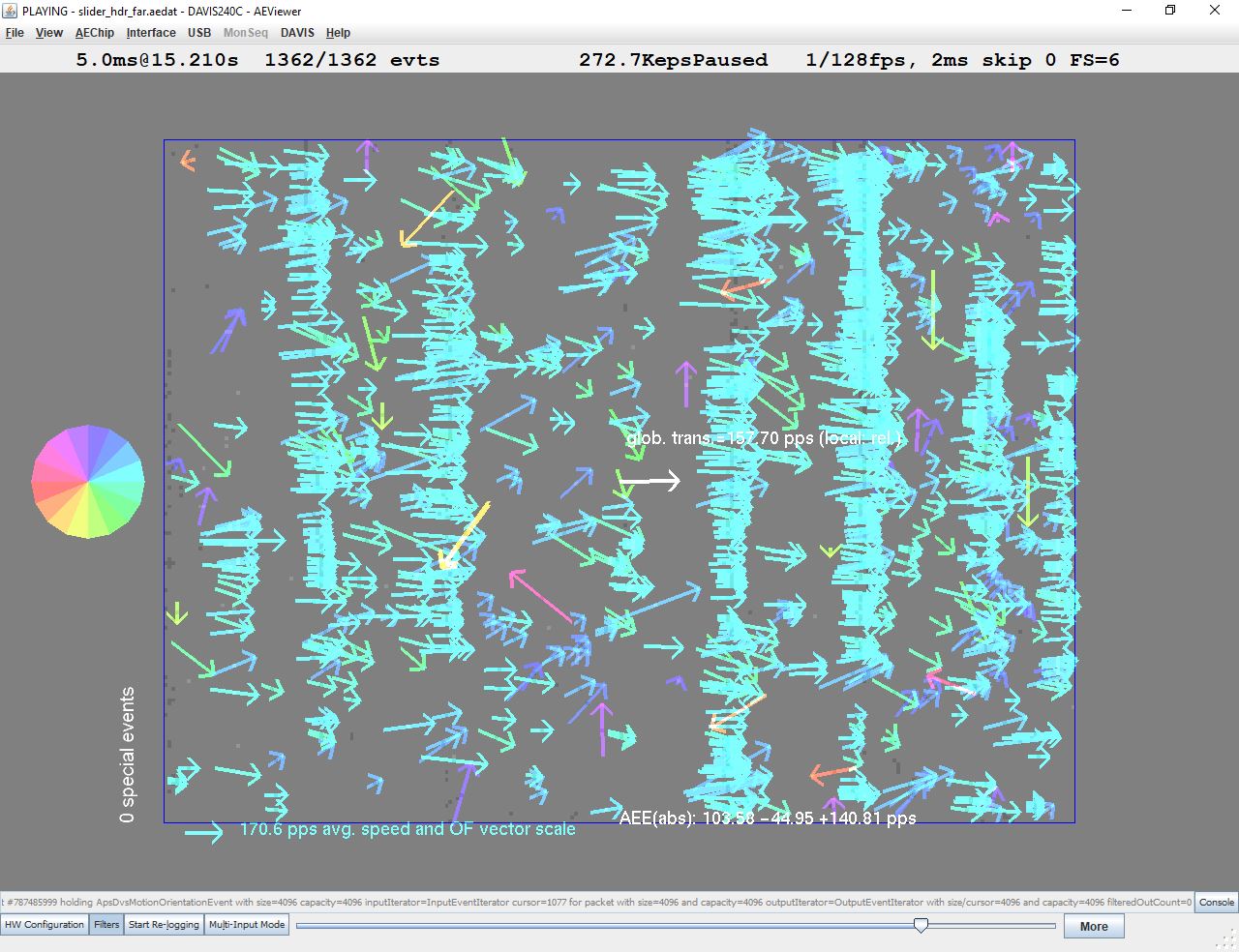}
        \end{center}
        \caption{LP}
        \label{fig:slider_hdr_far_f}
   \end{subfigure}
    \caption{Result of ABMOF, ABMOF\_LK and standard LK on image frames on \texttt{slider\_hdr\_fast}. For \ref{fig:slider_hdr_far_a} and \ref{fig:slider_hdr_far_b}, we use \textsl{AreaEventNumber} with feedback enabled, $r=4$, and $s=2$, 
    and used standard LK on successive APS images in \ref{fig:slider_hdr_far_c}.}
   \label{fig:slider_hdr_far}
\end{figure*}

\begin{figure*}[h!]
       \begin{subfigure}[b]{0.5\textwidth}
        \begin{center}
            \includegraphics[scale=0.9]{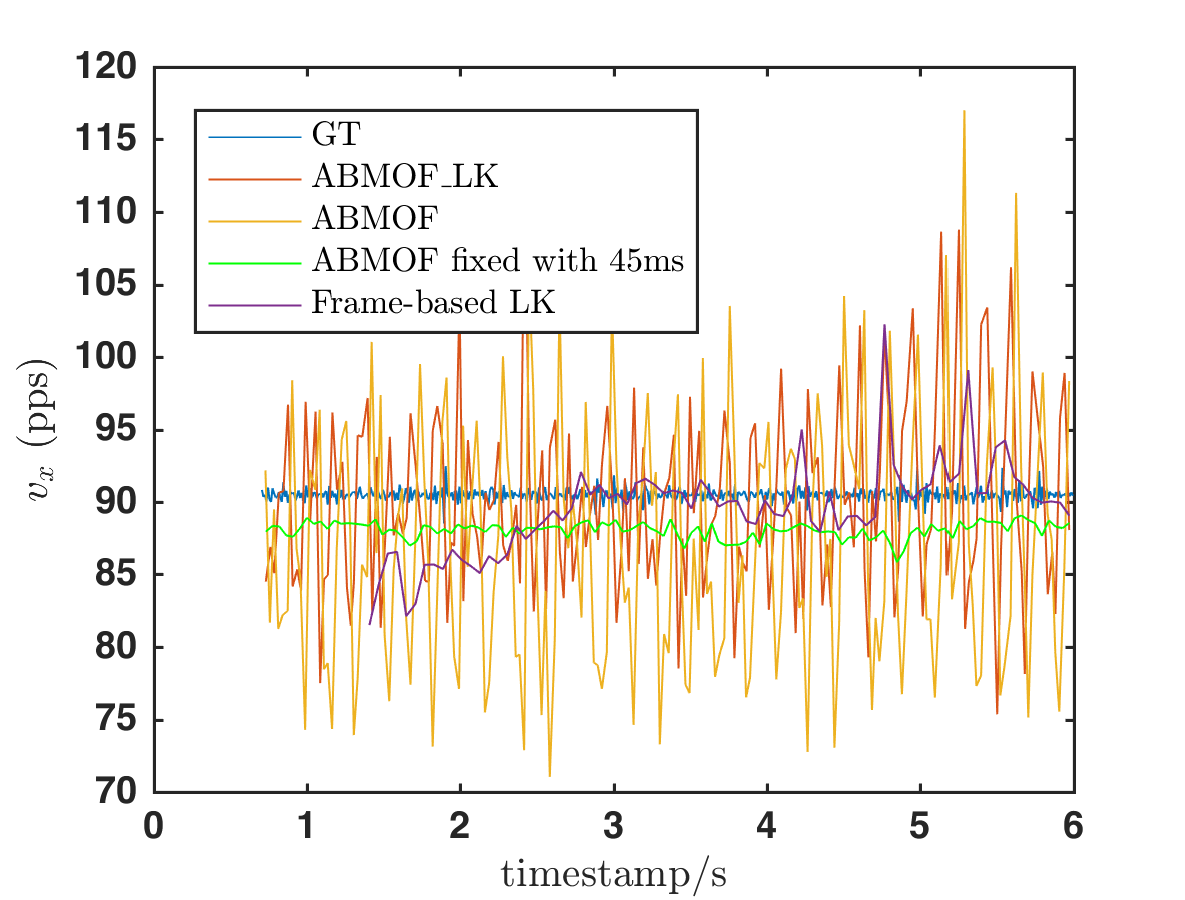}
        \end{center}
        \caption{$v_x$ for \texttt{slider\_hdr\_far}
        }
        \label{fig:sliderx}
    \end{subfigure}
    \begin{subfigure}[b]{0.5\textwidth}
        \begin{center}
            \includegraphics[scale=0.9]{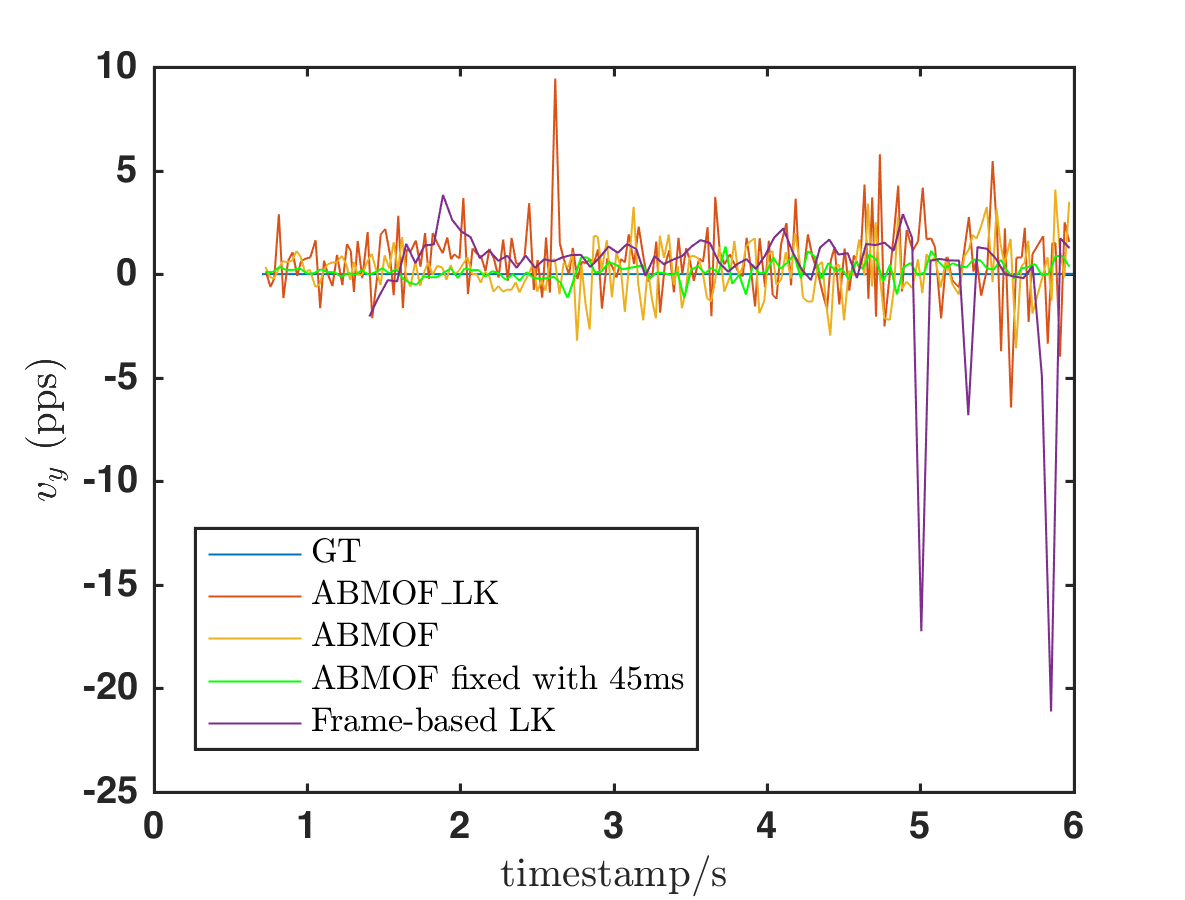}
        \end{center}
        \caption{$v_y$ for \texttt{slider\_hdr\_far}}
        \label{fig:slidery}
   \end{subfigure}
\caption{Comparison of measured and ground truth flow between OF methods
on \texttt{slider\_hdr\_far}}
\label{fig:LK_and_ABMOF}
\end{figure*}

Fig.~\ref{fig:slider_hdr_far} shows the qualitative results of \textrm{ABMOF} and \textrm{ABMOF\_LK} on the \texttt{slider\_hdr\_far} data. This is a high dynamic range scene of a flat poster with uniform flow about about 90\,pps. Because of the lighting contrast, the APS images are sometime extremely over- or underexposed, but the DVS events respond to the local brightness changes.
Table~\ref{tab:exp2_result} reports the quantitative comparison. 
By the VO groundtruth to OF groundtruth conversion, we can compare them over time, as shown in Fig.~\ref{fig:LK_and_ABMOF}. 
Table~\ref{tab:exp2_result} shows that 
ABMOF\_LK's GF error on the \texttt{slider\_hdr\_far} data is 
less than 1pps and ABMOF's GF error is less than 4pps.
\textrm{ABMOF\_LK} is more accurate than \textrm{ABMOF}, 
but has much lower ED. Fig.~\ref{fig:sliderx} 
shows a very clear periodic oscillation in $v_x$
for both \textsl{ABMOF} methods, which is caused
by the simple bang-bang control of the slice duration coupled with 
match distance quantization.
This oscillation is 
confirmed by the trace \textsl{ABMOF fixed with 45ms}, where we fixed $d=45$ms; its
$v_x$ flow is a bit too small because of the quantization of match distance. 
The conventional LK method on the frames also obtains
the average correct flow (and does not have the controller
oscillation), but as seen in Fig.~\ref{fig:slider_hdr_far_c},
this estimate is sometimes based on a single keypoint. 
That is the cause the 
outliers for \textsl{Frame\_based LK} 
in Fig.~\ref{fig:slidery} around 5s 
and 6s where the frame LK method suffered large aperture error. 

This experiment validates that the slice
rotation methods result in quantitative 
flow magnitude that is the same as from Frame based LK .
The ABMOF methods are oscillatory using the current $k$ controller, 
but have much higher density than the frame-based LK method.
All ABMOF methods are all much more accurate and less 
noisy than the prior DS, EBLK, and LP methods.

\begin{table*}[!htbp]
    \caption{Comparison of algorithm's overall accuracy on \texttt{slider\_hdr\_far}.}\label{tab:overall}
        \begin{center}
            \begin{tabular}{*7c}
                \toprule
                \bftab method & \bftab event density  & \bftab global flow (pps) & \bftab AEE (pps) & \bftab AAE ($^{\circ}$)\\[1ex]
                \midrule
                \textrm{Groundtruth} &  -  & $[90.50, 0]\pm[0.43, 0]$ & - & -\\[1ex]
                \textrm{ABMOF\_{LK}} &  $0.39\%$  & $[89.75, 0.44]\pm[6.30, 3.56]$ & 8.75$\pm$27.51 & 2.95$\pm$3.41\\[1ex]
                 \textrm{ABMOF} & $37.96\%$  & $[86.85, 0.17]\pm[8.46, 1.25]$ & 12.68$\pm$16.28 & 3.66$\pm$8.31\\[1ex]
                 \textrm{Frame\_based\_LK} & -  & $[89.51, 0.20]\pm[3.20 3.48]$ & 5.47$\pm$42.07 & 1.30$\pm$4.72\\[2ex]
                 \textrm{DS}~(\cite{delbruck2008frame,rueckauer2016evaluation})  & $49.86\%$  & $[74.97, 2.98]\pm[17.42, 4.79]$ & 57.71$\pm$53.31 & 21.46$\pm$39.13\\[1ex]
                 \textrm{EBLK}~(\cite{Benosman2012,rueckauer2016evaluation}) & $17.53\%$  & $[28.06, -0.11]\pm[4.09, 1.32]$ & 60.32$\pm$15.92 & 13.52$\pm$25.51\\[1ex] 
                 \textrm{LP}~(\cite{benosman2014event, rueckauer2016evaluation}) & $83.88\%$  & $[161.14, 11.69]\pm[8.67, 12.13]$ & 99.00$\pm$75.86 & 16.99$\pm$24.41\\[1ex]
                 
                \bottomrule  
            \end{tabular}
        \end{center}
        \label{tab:exp2_result}
\end{table*}

\subsubsection{\texttt{pavement\_fast} scene}
\label{sec:pavement_fast}
\begin{figure*}[!htpb]
	\begin{subfigure}[b]{0.33\textwidth}
        \begin{center}
            \includegraphics[width=6cm]{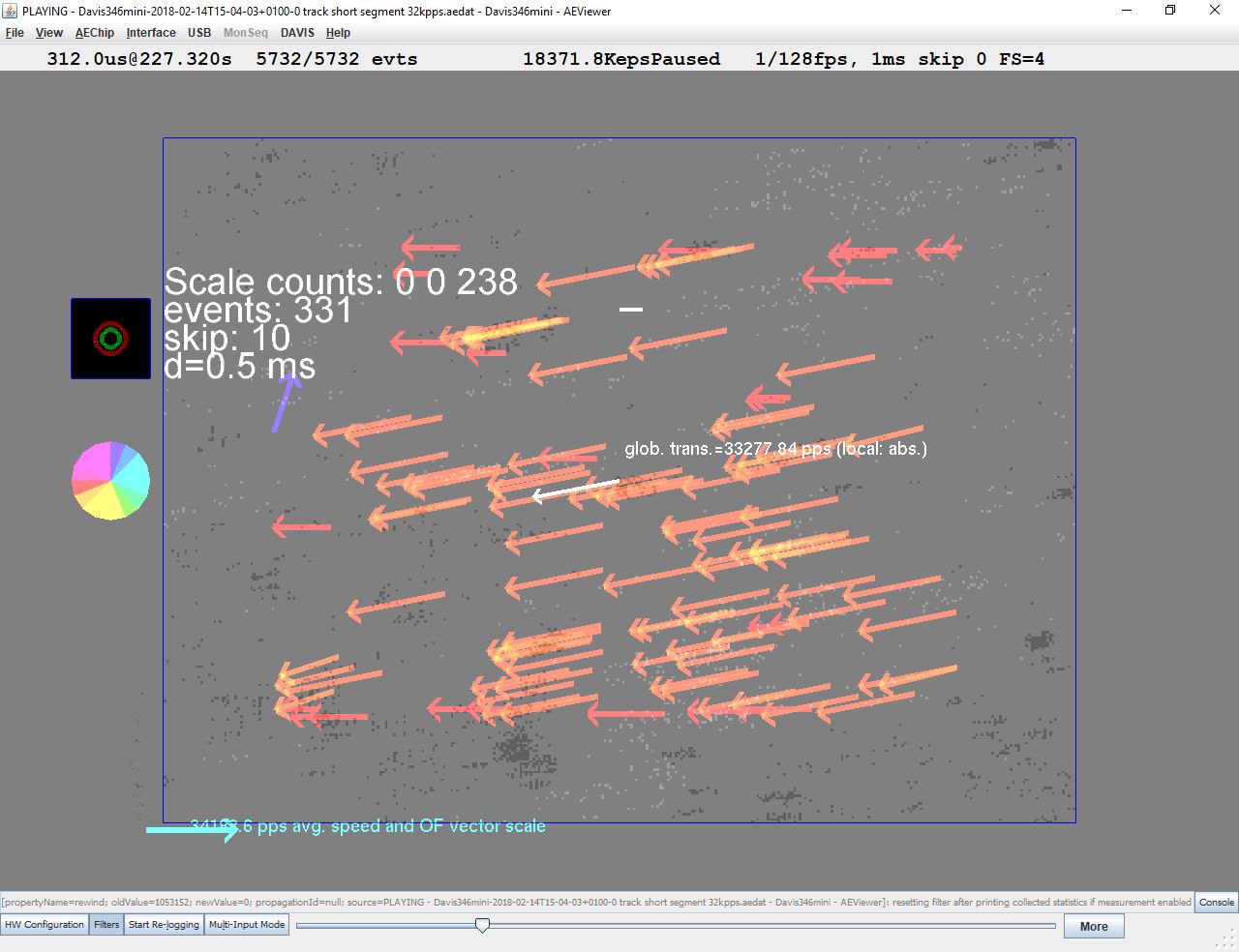}
        \end{center}
        \caption{\textrm{ABMOF}}
        \label{fig:speedo_a}
    \end{subfigure}
	\begin{subfigure}[b]{0.33\textwidth}
        \begin{center}
            \includegraphics[width=6cm]{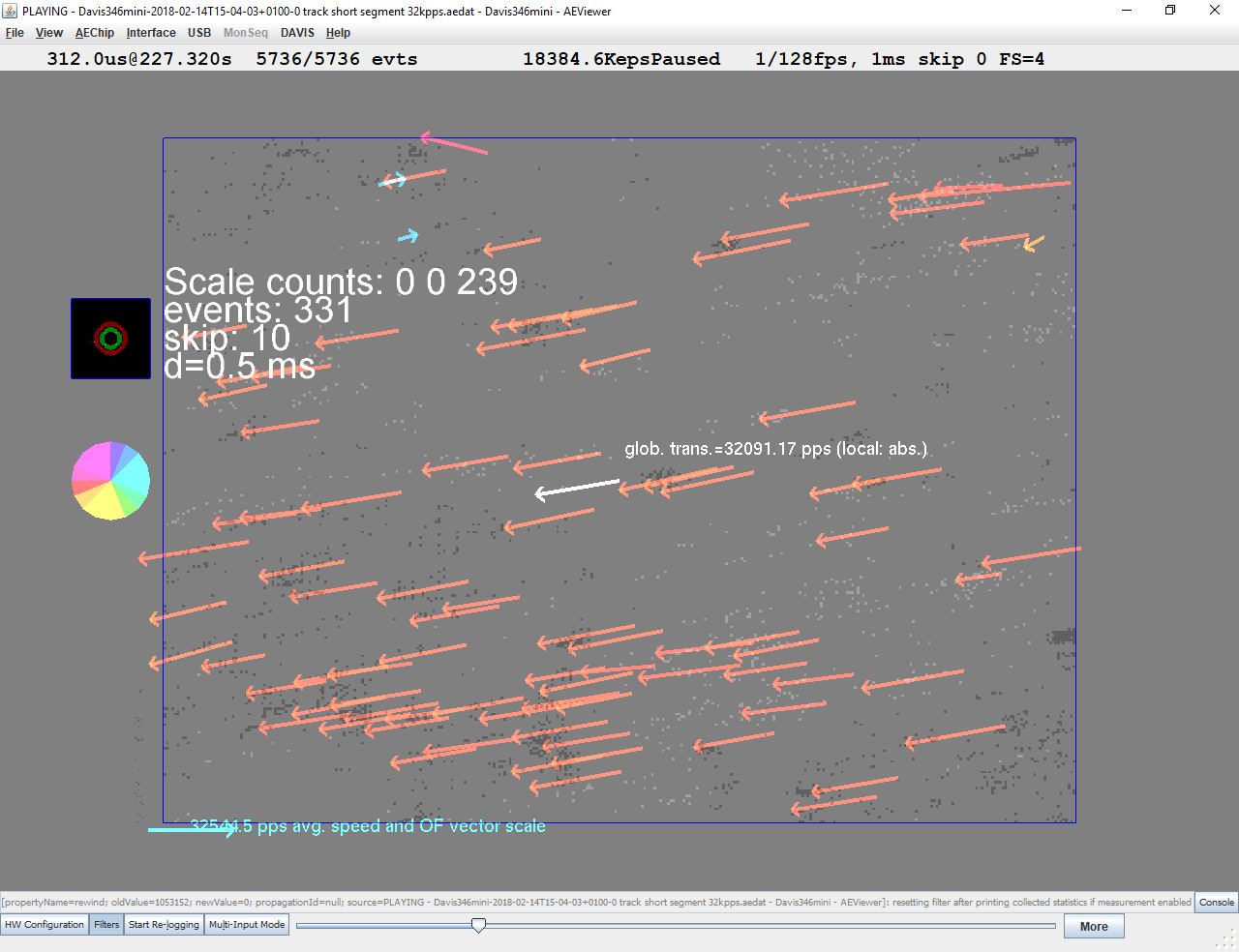}
        \end{center}
        \caption{ABMOF\_{LK}}
        \label{fig:speedo_b}
    \end{subfigure}
    \begin{subfigure}[b]{0.33\textwidth}
        \begin{center}
            \includegraphics[width=6cm]{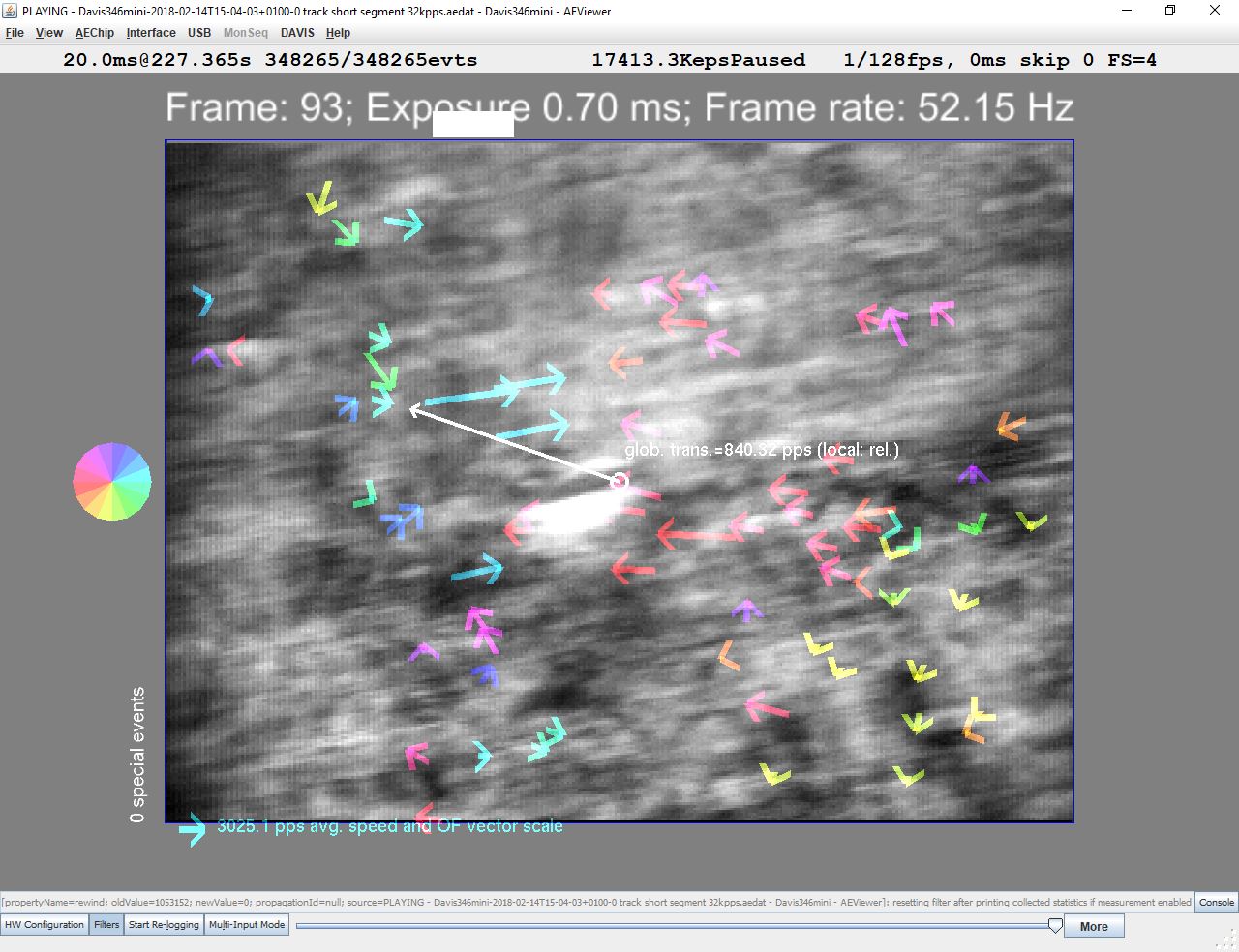}
        \end{center}
        \caption{APS image 2 and LK result}
        \label{fig:speedo_c}
   \end{subfigure}
    \caption{Result of ABMOF, ABMOF\_LK and standard LK on image frames on \texttt{pavement\_fast}. For \ref{fig:speedo_a} and \ref{fig:speedo_b}, we fixed $d=450$\,us, $r=12$, and $s=3$, 
    and used standard LK on successive APS images in \ref{fig:speedo_c}.}
   \label{fig:speedo}
\end{figure*}

Fig.~\ref{fig:speedo}  shows the results of a very high speed experiment
on \texttt{pavement\_fast}, which was recorded from a car with a down-looking
camera aimed at the asphalt pavement road surface. The global flow is 
an extremely fast 32k\,pps, which means that a pixel crosses 
the 346-pixel array in about 10\,ms. 
Figs.~\ref{fig:speedo_a}~-~\ref{fig:speedo_b} 
compares ABMOF and ABMOF\_LK on DVS time slices, and Fig.~\ref{fig:speedo_c} shows conventional LK 
on successive DAVIS APS images (the 2nd image is shown under the flow result). 
Both ABMOF and ABMOF\_LK correctly measure the true flow using a slice duration of only 450\,us, 
equivalent to
a frame rate of 22\,kHz and a 14\,pixel displacement between slices.
The consecutive APS image frames were 
collected at the maximum frame rate of 50\,Hz, but because
the motion is so fast, even the short DAVIS 
global shutter exposure of 0.7\,ms resulted in visible image blur of several pixels. 
And since the consecutive frames are separated by 20ms, 
the images are completely uncorrelated 
and the resulting flow is meaningless as seen in Fig.~\ref{fig:speedo_c}. 

\begin{figure*}[!htpb]
	\begin{subfigure}[b]{6cm}
        \begin{center}
            \includegraphics[width=6cm]{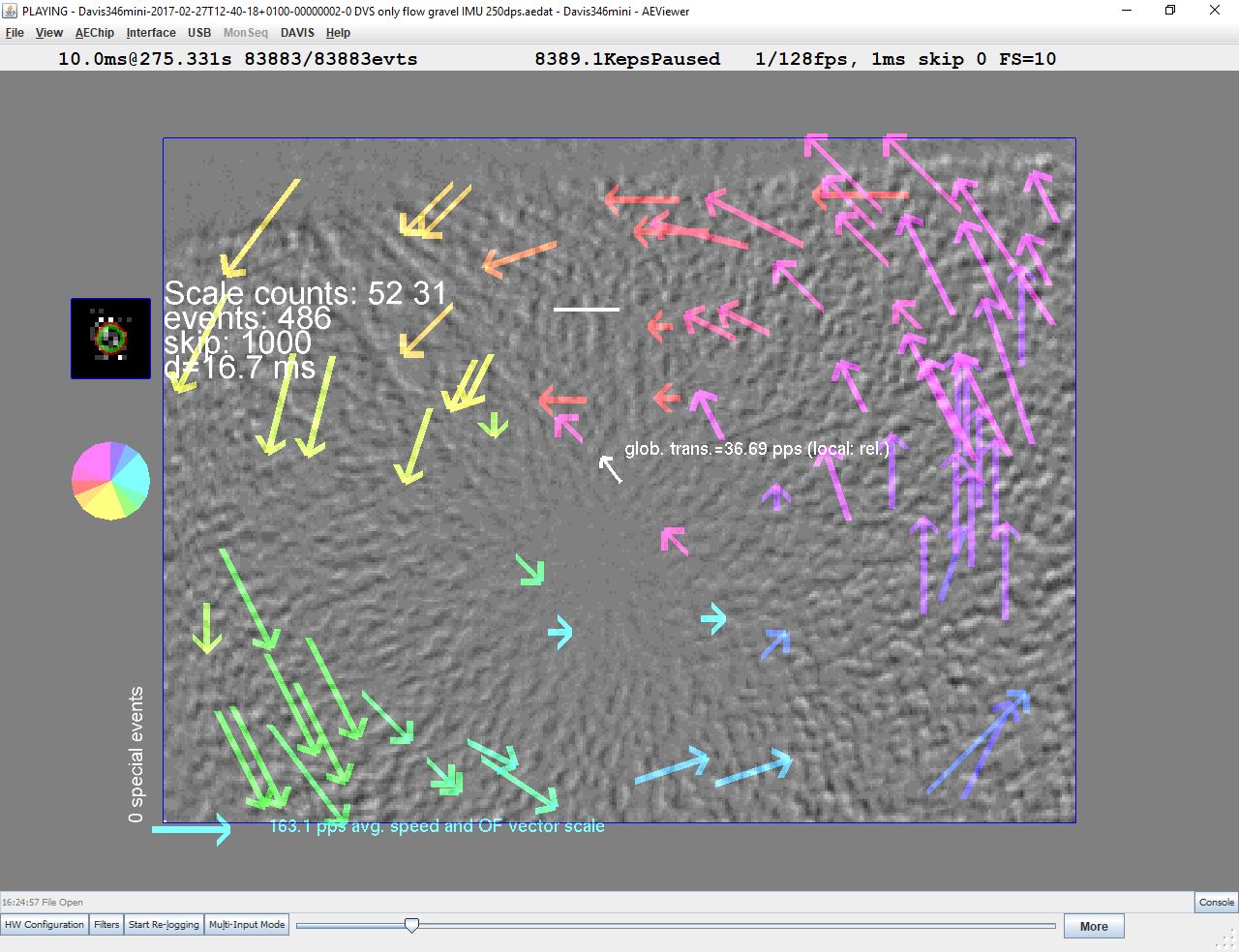}
        \end{center}
        \caption{\textrm{ABMOF} for \texttt{gravel}}
        \label{fig:boxes}
    \end{subfigure}
	\begin{subfigure}[b]{6cm}
        \begin{center}
            \includegraphics[width=6cm]{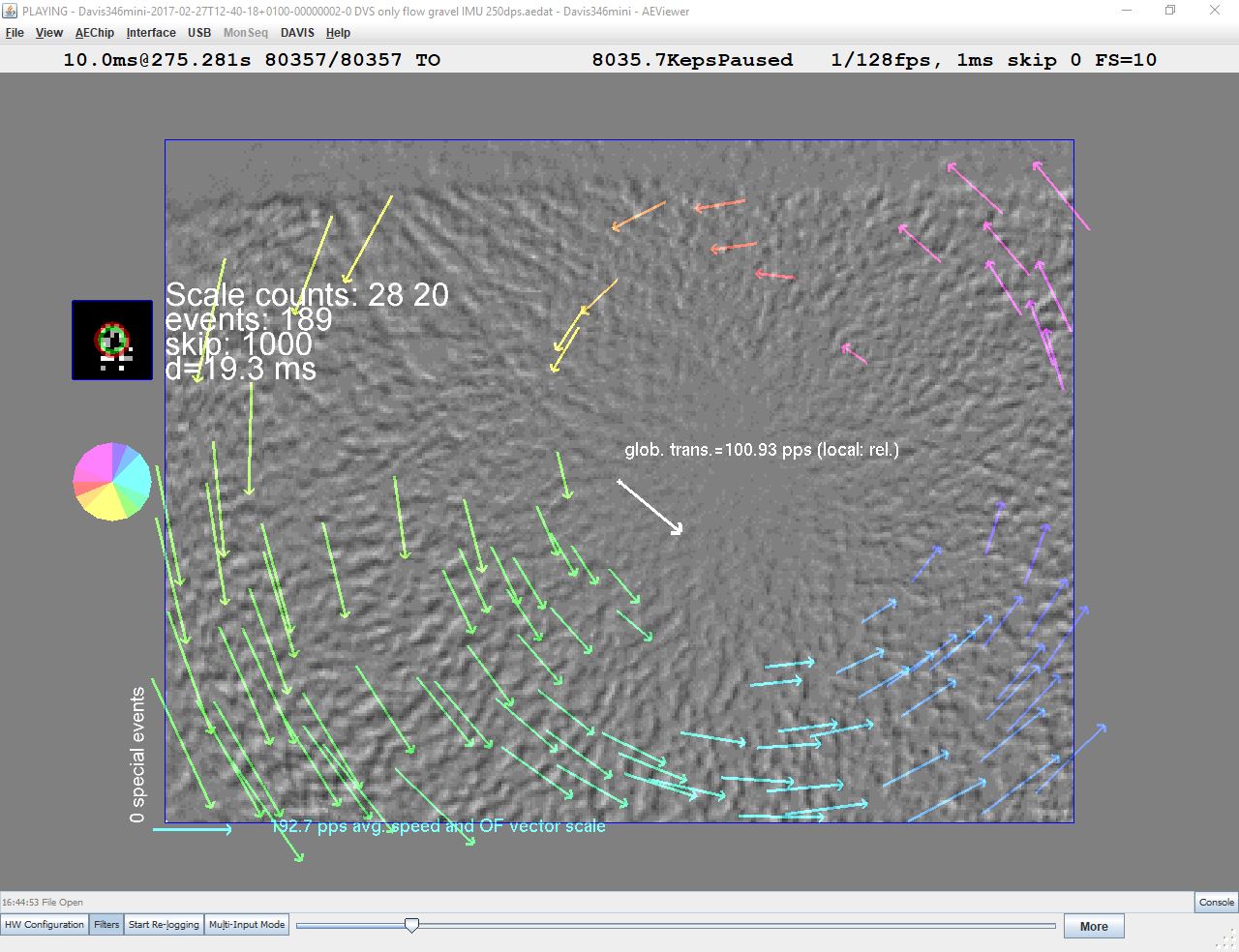}
        \end{center}
        \caption{$ABMOF_{LK}$ for \texttt{gravel}}
        \label{fig:boxes}
    \end{subfigure}
    \begin{subfigure}[b]{6cm}
        \begin{center}
            \includegraphics[width=6cm]{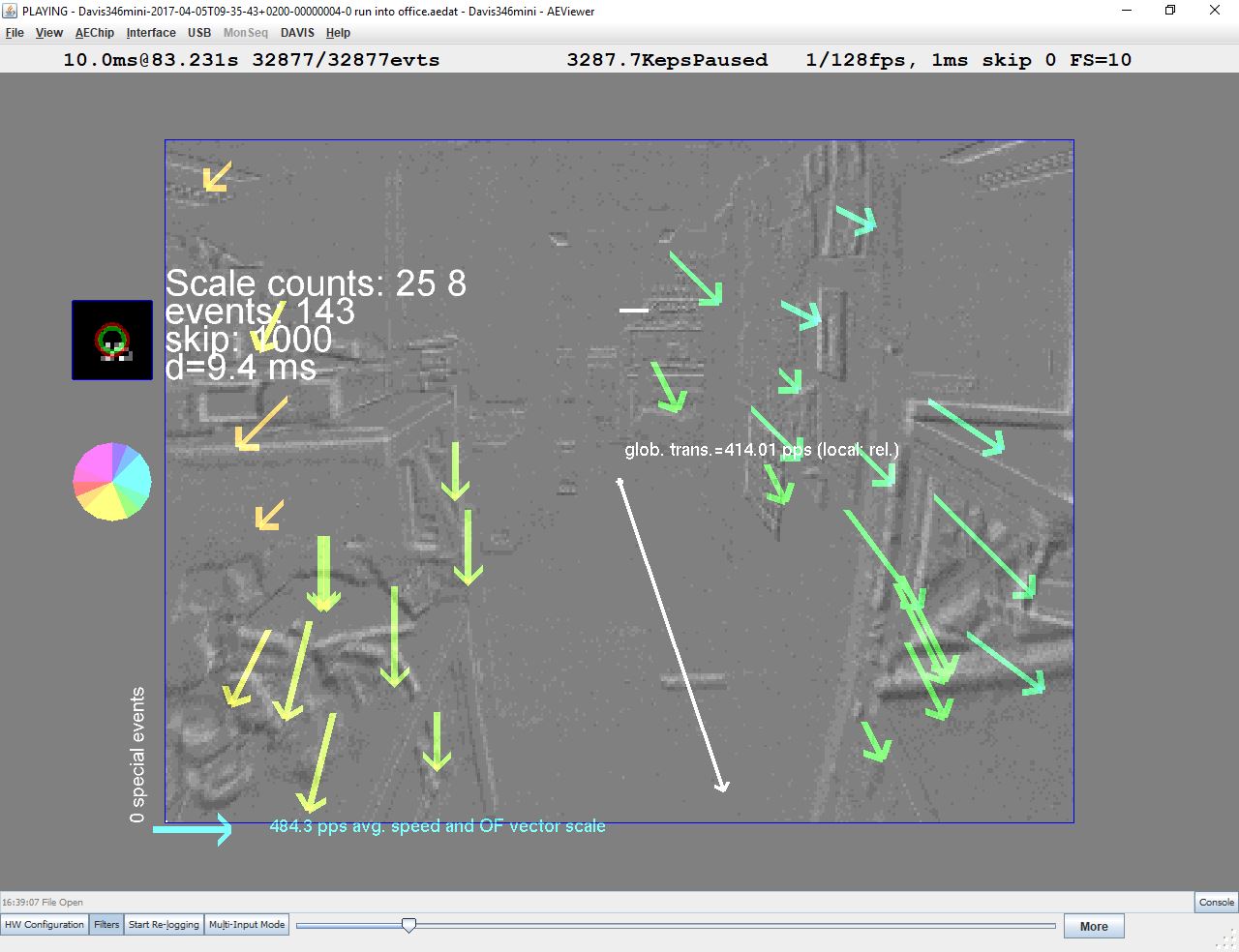}
        \end{center}
        \caption{\textrm{ABMOF} for \texttt{office}}
        \label{fig:grassPavement}
   \end{subfigure}
    \begin{subfigure}[b]{6cm}
        \begin{center}
            \includegraphics[width=6cm]{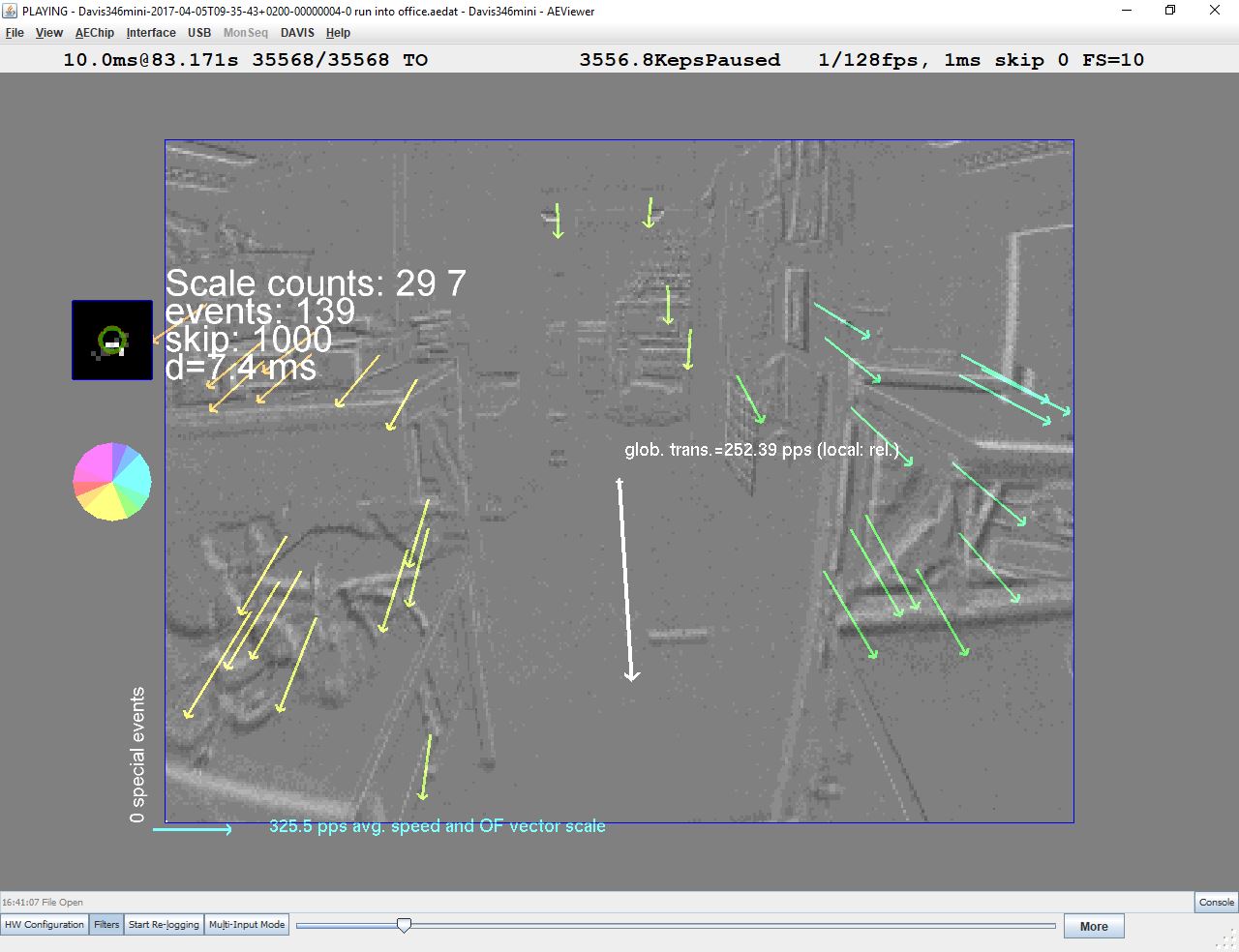}
        \end{center}
        \caption{\textrm{ABMOF\_{LK}} for \texttt{office}}
        \label{fig:grassPavement}
   \end{subfigure}
   \begin{subfigure}[b]{6cm}
        \begin{center}
            \includegraphics[width=6cm]{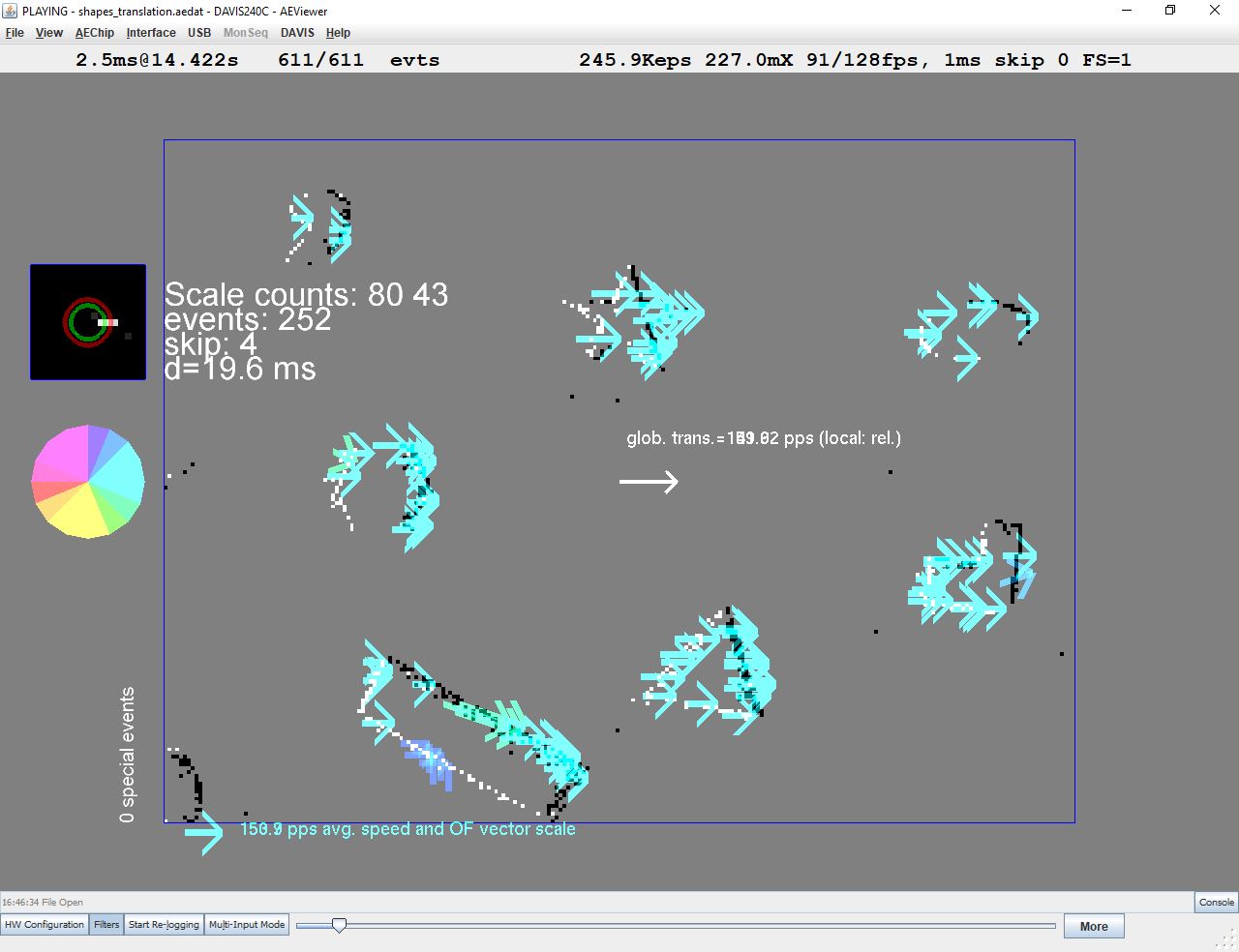}
        \end{center}
        \caption{\textrm{ABMOF} for \texttt{shapes}}
        \label{fig:shape}
   \end{subfigure}
   \begin{subfigure}[b]{6cm}
        \begin{center}
            \includegraphics[width=6cm]{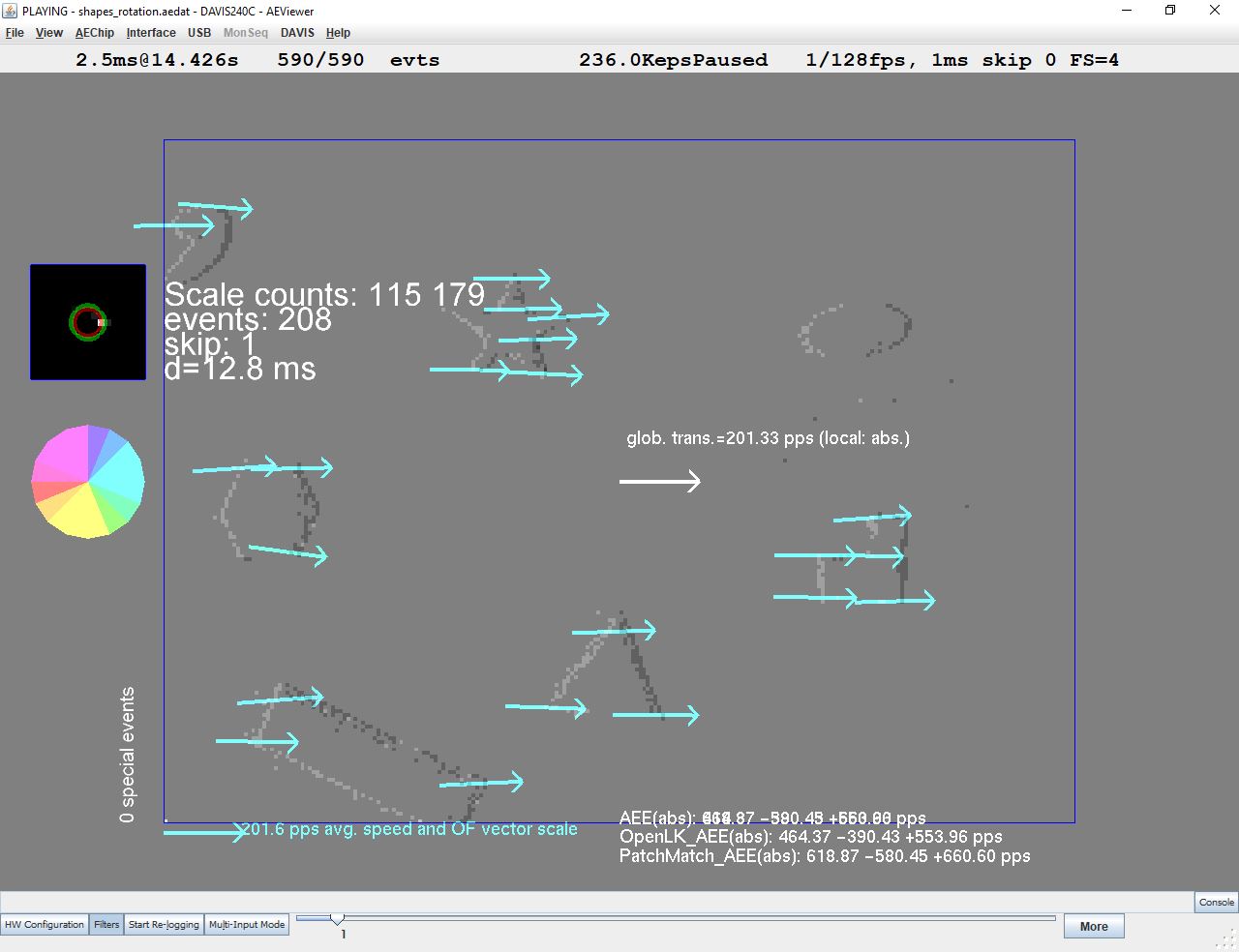}
        \end{center}
        \caption{\textrm{ABMOF\_LK} for \texttt{shapes}}
        \label{fig:shape_LK}
   \end{subfigure}
    \caption{Result of the algorithms in different scenes. All scenes were captured using identical $s=2$ scales, block size $b=21$ pixels, using diamond search, with search distance $r= 4$ pixels and using feedback control of \textsl{AreaEventNumber} $k$. OF color and angle represents direction according to the color wheel and vectors' length means OF speed relative to the scale shown at bottom left of each frame. The histogram above each color wheel shows the OF distribution and mean match distance (green circle). The white arrow from center of image shows global average flow. The white statistics text shows the number of OF events for each scale, the number of OF events, the current skip count $p$, and the last slice duration $d$.}
   \label{fig:result}
\end{figure*}

\subsection{Type II experiment result} \label{Type II}
The final experiments are from natural scenes that contain a range of directions and speeds. Since we lack ground truth OF for these natural scenes, 
we only show the qualitative comparison of ABMOF and ABMOF\_LK. 
These results are shown in Fig~\ref{fig:result}.
We use vectors to represent the OF result; color also shows the direction. 
For clarity, we set $p=1000$ for ABMOF.
The ABMOF and ABMOF\_LK  produce very similar OF for these natural scenes.
For \texttt{shapes}, ABMOF flow is quite dense along object edges and the large block size of 21\,pixels results in true OF rather than normal flow. 
For this same scene, ABMOF\_LK attaches OF only to object corners; 
ABMOF\_LK misses the OF on the upper right ellipse, 
but the overall flow is more uniform. 

\section{Conclusion and Discussion}
\label{sec:conclusion}

We proposed improvements for the basic BMOF algorithm for DVS.
A new pipeline adjusts the slice duration based on the local movement rather than the global motion. 
Using multi-scale bitmaps allows a larger range of movement 
speeds to be economically computed and makes the operation more robust to noisy sensor data.
A feedback mechanism for slice duration 
makes the average displacement between two slices close to the half of the search distance. 
These improvements on the basic BMOF achieves a 
good trade off between good quality of optical flow estimation and a low computation cost. 
We also compared our generated adaptive slices with the constant time duration slices. 

The reason we implemented the \textrm{ABMOF\_LK} 
and show the results here is to show the event slices 
generated by our adaptive method are robust to different 
scenes and can provide better grayscale images for 
frame-based algorithms to process.
Although \textrm{ABMOF\_LK} clearly produces more accurate OF, 
it is a frame-based OF method that 
must process every pixel in each frame and produces very sparse output. 
The gradient-based LK algorithm is also much more 
difficult and expensive to implement in logic circuits compared with \textrm{ABMOF}.

The results shows that the accuracy of the algorithm mainly 
depends on the quality of the generated slices. 
By using the rather large block dimension $r=21$, 
ABMOF avoids most aperture problems except on 
extended edges longer than the block dimension. 

The dynamic range of speeds allowed by ABMOF is determined by the search
distance $r$ and the number of scales $s$. In any one moment, the range of
match distances spans from 0 to $r 2^s$ pixels along each axis, e.g., with
search distance $r=4$ and $s=3$ scales 
the OF can span 0 to $\pm 32$ pixels, although the speed
resolution decreases as the scale increases. In our experience this is
sufficient range to cover real scenes where the camera
is rotating, or translating through a cluttered environment with nearby and
far objects. With the adaptive slice duration, fast motion can result in slice durations that are 
fractions of a millisecond, as in the \texttt{pavement\_fast} 
example of Sec~\ref{sec:pavement_fast}, allowing measurement
of speeds $>10$k\,pps. This result was previously only 
the domain of high-end gaming mouse sensors
such as~\cite{pixart_imaging_inc._pmw3360dm-t2qu:_2014}; 
these are
capable of up to several thousand FPS 
but require active illumination and have less than $50x50$ pixel
arrays that are more than a $100$ times fewer pixels than the DAVIS used here; 
also, they only measure global translational flow.

By extrapolating the FPGA hardware implementation costs from \cite{liu2017block}, 
we estimate that ABMOF can be implemented on a medium sized FPGA fabric, 
such as Xilinx Zynq-7000 family chip XC7Z100. The total/percent utilization
will require about 35k/6\% Flip-Flops, 100k/36\%
Look-Up Tables and 400\,kb/1\% block RAM. 
In addition, the design from \cite{liu2017block} is not optimized.
The resulting IP block could later be 
integrated together with the sensor in a custom digital core. 

The most widely used applications of OF are in optical mice and video compression, 
where probably at least a billion ICs have 
been produced that estimate motion based on block matching. 
In robotics, most visual odometry (\textbf{VO}) pipelines do not currently use OF, 
but an economical implementation could enable 
direct VO based on DVS in hardware, 
rather than the impressive but
 expensive software solutions~\cite{rebecq2017real,zhu_event-based_2017,vidal_ultimate_2018}.
Recent success in 
combining DVS with convolutional neural 
networks (CNNs) by  
using constant event number 
frames~\cite{moeys_steering_2016,lungu_live_2017,amir_low_2017,fischl_neuromorphic_2017}
also can benefit from the smarter ABMOF slice methods, and the OF could provide useful 
input channel information to better 
enable dynamic scene analysis.

\appendices
 
 \ifCLASSOPTIONcaptionsoff
   \newpage
 \fi
 
\bibliographystyle{IEEEtran}
 \bibliography{main}
 
\end{document}